\documentclass[aps,pre,notitlepage]{revtex4-1}

\usepackage{graphicx}
\usepackage{float}
\usepackage{placeins}
\usepackage{color, colortbl}
\usepackage{multirow}
\usepackage{array}
\usepackage[overload]{empheq}
\usepackage{amsmath,amssymb,amsfonts}
\usepackage{subfigure}
\usepackage[top=0.85in,left=1.8in,footskip=0.75in]{geometry}

\newcolumntype{M}[1]{>{\centering\arraybackslash}m{#1}}

{
\setlength{\tabcolsep}{3pt}

\begin{document}

\title{Clustering Algorithms: A Comparative Approach}

\author{Mayra Z. Rodriguez\textsuperscript{1}}
\author{Cesar H. Comin\textsuperscript{2}}
\email{chcomin@gmail.com}
\author{Dalcimar Casanova\textsuperscript{3}}
\author{Odemir M. Bruno\textsuperscript{2}}
\author{Diego R. Amancio\textsuperscript{1}}
\email{diego.raphael@gmail.com}
\author{Francisco A. Rodrigues\textsuperscript{1}}
\author{Luciano da F. Costa\textsuperscript{2}}

% %%
\affiliation{\textbf{1} Institute of Mathematics and Computer Science, University of S\~{a}o Paulo, S\~{a}o Carlos, S\~ao Paulo, Brazil \\
			\textbf{2} S\~{a}o Carlos Institute of Physics, University of S\~{a}o Paulo, PO Box 369, S\~{a}o Carlos, SP, Brazil  \\
			\textbf{3} Federal University of Technology - Paran\'a, Paran\'a, Brazil}

%\date{\today}

\begin{abstract}
Many real-world systems can be studied in terms of pattern recognition tasks, so that proper use (and understanding) of machine learning methods in practical applications becomes essential. While a myriad of classification methods have been proposed, there is no consensus on which methods are more suitable for a given dataset. As a consequence, it is important to comprehensively compare methods in many possible scenarios. In this context, we performed a systematic comparison of 7 well-known clustering methods available in the R language. In order to account for the many possible variations of data, we considered artificial datasets with several tunable properties (number of classes, separation between classes, etc). In addition, we also evaluated the sensitivity of the clustering methods with regard to their parameters configuration. The results revealed that, when considering the default configurations of the adopted methods, the spectral approach usually outperformed the other clustering algorithms. We also found that the default configuration of the adopted implementations was not accurate. In these cases, a simple approach based on random selection of parameters values proved to be a good alternative to improve the performance. All in all, the reported approach provides subsidies guiding the choice of clustering algorithms.
\end{abstract}

\maketitle

\section{Introduction}

In recent years, the automation of data collection and recording implied a deluge of information about many different kinds of systems~\cite{Golder1878,Michel176,10.1371/journal.pone.0006022,Amancio2012427,Dean:2008:MSD:1327452.1327492,Viana2013371,Aggarwal,Ridgeway2003}. As a consequence, many methodologies aimed at organizing and modeling data have been developed~\cite{fayyad1996data}. Such methodologies are motivated by their widespread application in diagnosis~\cite{bellazzi2008predictive}, education~\cite{abdullah2011extracting}, forecasting~\cite{khashei2010artificial}, and many other domains~\cite{joachims1998text}. The definition, evaluation and application of these methodologies are all part of the machine learning field~\cite{witten2005data}, which became a major subarea of computer science and statistics due to their crucial role in the modern world. 

Machine learning encompasses different topics such as regression analysis~\cite{wang2010high}, feature selection methods~\cite{blum1997selection}, and classification~\cite{witten2005data}.  The latter involves assigning classes to the objects in a dataset. Three main approaches can be considered for classification: supervised, semi-supervised and unsupervised  classification. In the former case, the classes, or labels, of some objects are known beforehand, defining the training set, and an algorithm is used to obtain the classification criteria. 
Semi-supervised classification deals with training the algorithm using both labeled and unlabeled data. They are commonly used when manually labeling a dataset becomes costly. Lastly, 
unsupervised classification, henceforth referred as \emph{clustering}, deals with \emph{defining} classes from the data without knowledge of the class labels. The purpose of clustering algorithms is to identify groups of objects, or clusters, that are more similar to each other than to other clusters. Such an approach to data analysis is closely related to the task of creating a model of the data, that is, defining a simplified set of properties that can provide intuitive explanation about relevant aspects of a dataset. Clustering methods are generally more demanding than supervised approaches, but provide more insights about complex data. This type of classifiers constitute the main object of the current work.  

Because clustering algorithms involve several parameters, often operate in high dimensional spaces, and have to cope with noisy, incomplete and sampled data, their performance can vary substantially for different applications and types of data. For such reasons, several different approaches to clustering have been proposed in the literature (e.g. ~\cite{jing2007entropy,suzuki2006pvclust,camastra2005novel}).  In practice, it becomes a difficult endeavor, given a dataset or problem, to choose a suitable clustering approach.  Nevertheless, much can be learned by comparing different clustering methods.  Several previous efforts for comparing clustering algorithms have been reported in the literature~\cite{jung2014clustering,kinnunen2011comparison, abbas2008comparisons,pirim2012clustering,COSTA2004,de2008clustering,dougherty2002inference,Brohee2006, maulik2002performance, fraley1998many}. Here, we focus on generating a diversified and comprehensive set of artificial data containing not only distinct number of classes, features, number of objects and separation between classes, but also a varied structure of the involved groups (e.g. possessing predefined correlation distributions between features).  The purpose of using artificial data is the possibility to obtain an unlimited number of samples and to systematically change any of the aforementioned properties of a dataset. Such features allow the clustering algorithms to be comprehensive and strictly evaluated in a vast number of circumstances, and also grants the possibility of quantifying the sensitivity of the performance with respect to small changes in the data.   

Here we associate performance with the similarity between the known labels of the objects and those found by the algorithm. Many measurements have been defined for quantifying such similarity~\cite{halkidi2001clustering}, we compare the Jaccard index~\cite{citeulike:7707101}, Adjusted Rand index~\cite{Hubert1985}, Fowlkes-Mallows index~\cite{Fowlkes83} and Normalized mutual information~\cite{Strehl02clusterensembles}.  A modified version of the procedure developed by~\cite{hirschberger2007randomly} was used to create 270 distinct datasets, which were used in order to quantify the performance of the clustering algorithms. In Section~\ref{s:artfData} we describe the adopted procedure and the respective parameters used for data generation. Related approaches include~\cite{amancio2014systematic}. 

Each clustering algorithm relies on a set of parameters that needs to be adjusted in order to achieve viable performance, which corresponds to an important point to be addressed while comparing clustering algorithms. A long standing problem in machine learning is the definition of a proper procedure for setting the parameter values~\cite{berkhin2006survey}. In principle, one can apply an optimization procedure (e.g., simulated annealing~\cite{hwang1988simulated} or genetic algorithms~\cite{goldberg1988genetic}) to find the parameter configuration providing the best performance of a given algorithm. Nevertheless, there are two major problems with such an approach. First, adjusting parameters to a given dataset may lead to overfitting~\cite{hawkins2004problem}. That is, the specific values found to provide good performance may lead to worse results if new data is considered. Second, parameter optimization can be unfeasible in some cases, given the time complexity of many algorithms, combined with their typically large number of parameters. Ultimately, many researchers resort to applying classifier or clustering algorithms using the default parameters provided by the software.  Therefore, efforts are required for evaluating and comparing the performance of clustering algorithms in the optimization and default situations.  In the following, we consider some representative examples of algorithms applied in the literature~\cite{berkhin2006survey, jain1999data}. 

Clustering algorithms have been implemented in several programming languages and packages. During the development and implementation of such codes, it is common to implement changes or optimizations, leading to new versions of the original methods. The current work focuses on the comparative analysis of several clustering algorithm found in popular packages available in the R programming language~\cite{manualR}. This choice was motivated by the popularity of the R language in the data mining field, and by virtue of the well-established clustering packages it contains. This study is intended to assist researchers who have programming skills in R language, but with little experience in clustering of data.

The algorithms are evaluated on three distinct situations. First, we consider their performance when using the default parameters provided by the packages. Then, we consider the performance variation when single parameters of the algorithms are changed, while the rest are kept at their default values. Finally, we consider the simultaneous variation of all parameters by means of a random sampling procedure. We compare the results obtained for the latter two situations with those achieved by the default parameters, in such a way as to investigate the possible improvements in performance which could be achieved by modifying the algorithms.

The text is divided as follows. We start by revising some of the main approaches to clustering algorithms comparison.  Next, we describe the clustering methods considered in the analysis, we also present the R packages implementing such methods. In Section 4 we detail the data generation method and the performance measurements used to compare the algorithms. In Section 5 we present the performance results obtained for the default parameters (Section 5.1), for single parameter variation (Section 5.2) and for random parameter sampling (Section 5.3).

\section{Related works}

Previous approaches for comparing the performance of clustering algorithms can be divided according to the nature of used datasets. While some studies use either real-world or artificial data, others employ both types of datasets to compare the performance of several clustering methods.
%
%in three large groups, according to the use of real world, artificial respectively to the use of (i) real world data, (ii) artificial data, and (iii) a mixture of artificial and real world data.

A comparative analysis using real world dataset is presented in several works~\cite{kou2014evaluation,COSTA2004,
erman2006traffic,de2008clustering,kinnunen2011comparison,jung2014clustering}. Some of these works are reviewed briefly in the following.
In~\cite{kou2014evaluation}, the authors propose an evaluation approach based in a multiple criteria decision making in the domain of financial risk analysis over three real world credit risk and bankruptcy risk datasets. More specifically, clustering algorithms are evaluated in terms of a combination of clustering measurements, which includes a collection of external and internal validity indexes. Their results show that no algorithm can achieve the best performance on all measurements for any dataset and, for this reason, it is mandatory to use more than one performance measure to evaluate clustering algorithms. 

In~\cite{kinnunen2011comparison}, a comparative analysis of clustering methods was performed in the context of text-independent speaker verification task, using three dataset of documents. Two approaches were considered: clustering algorithms focused in minimizing a distance based objective function and a Gaussian models-based approach. The following algorithms were compared: k-means, random swap, expectation-maximization,  hierarchical clustering, self-organized maps (SOM) and fuzzy c-means. The authors found that the most important factor for the success of the algorithms is the model order, which represents the number of centroid or Gaussian
components (for Gaussian models-based approaches) considered. Overall, the recognition accuracy was similar for clustering algorithms focused in minimizing a distance based objective function. When the number of clusters was small, SOM and hierarchical methods provided significantly poorer accuracy than the other methods. Finally, a comparison of the computational efficiency of the methods revealed that the split hierarchical method is the fastest clustering algorithm in the considered dataset.

In~\cite{de2008clustering}, five clustering methods were studied: k-means, multivariate Gaussian mixture, hierarchical clustering, spectral and nearest neighbor methods. Four proximity measures were used in the experiments: pearson and spearman correlation coefficient, cosine similarity and the euclidean distance. The algorithms were evaluated in the context of 35 gene expression data from either Affymetrix or cDNA chip platforms, using the adjusted rand index for performance evaluation. The multivariate Gaussian mixture method provided the best performance in recovering the actual number of clusters of the datasets. The k-means method displayed similar performance. In this same analysis, the hierarchical method led to limited performance, while the spectral method showed to be quite sensitive to the proximity measure employed.

In~\cite{COSTA2004}, experiments were performed to compare five different types of clustering algorithms: CLICK, self organized mapping-based method (SOM), k-means, hierarchical and dynamical clustering. Data sets of gene expression time series of the \emph{Saccharomyces cerevisiae} yeast were used. A k-fold cross-validation procedure was considered to compare different algorithms. 
The authors found that k-means, dynamical clustering and SOM obtained high accuracy in all experiments. On the other hand, hierarchical clustering presented a poor performance in clustering larger datasets, yielding low accuracy in some experiments.

A comparative analysis using artificial data is presented in~\cite{mingoti2006comparing, mangiameli1996comparison,parsons2004evaluating}.
In~\cite{parsons2004evaluating}, two subspace clustering methods were compared: MAFIA~(Adaptive Grids for Clustering Massive Data Sets)~\cite{Burdick:2001:MMF:645484.656386} and FINDIT~(A fast and intelligent subspace clustering algorithm using dimension voting)~\cite{parsons2004subspace}. The artificial data, modeled according to a normal distribution, allowed the control of the number of dimensions and instances. The methods were  evaluated in terms of both scalability and accuracy. In the former, the running time of both algorithms were compared for different number of instances and features. In addition, the authors assessed the ability of the methods in finding adequate subspaces for each cluster. They found that MAFIA discovered all relevant clusters, but one significant dimension was left out in most cases. Conversely, the FINDIT method performed better in the task of identifying the most relevant dimensions.  Both algorithms were found to scale linearly with the number of instances, however MAFIA outperformed FINDIT in most of the tests.

Another common approach for comparing clustering algorithms considers using a mixture of real world and artificial data (e.g. ~\cite{verma2003comparison, maulik2002performance, pirim2012clustering,  dougherty2002inference, Brohee2006}). In~\cite{maulik2002performance}, the performance of   k-means, single linkage and simulated annealing~(SA) was evaluated, considering different partitions obtained by validation indexes. The authors used two real world datasets obtained from \cite{dataset:UCI} and three artificial datasets~(having two dimensions and 10 clusters). The authors proposed a new  validation index called $I$ index that measures the  separation based on the maximum distance between clusters and compactness based on the sum of distances between objects and their respective centroids. They found that such an index was the most reliable among other considered indices, reaching its maximum value when the number of clusters is properly chosen.  In addition, the authors concluded that the studied SA-based algorithms outperformed the traditional k-means.

A systematic quantitative evaluation of four graph-based clustering methods was performed in~\cite{Brohee2006}. The compared methods were: markov clustering (MCL), restricted neighborhood search clustering (RNSC), super paramagnetic clustering (SPC), and molecular complex detection (MCODE). Six datasets modeling protein interactions in the \textit{Saccharomyces cerevisiae} and 84 random graphs were used for the comparison. For each algorithm, the robustness of the methods was measured in a twofold fashion: the variation of performance was quantified in terms of changes in the (i) methods parameters and (ii) dataset properties. In the latter, connections were included and removed to reflect uncertainties in the relationship between proteins. The restricted neighborhood search clustering method turned out to be remarkably robust to variations in the choice of method parameters, whereas the other algorithms were found to be more robust to dataset alterations.

\section{Clustering methods}

Many different types of clustering methods have been proposed in the literature~\cite{Guha98,Aggarwal2013,Karypis99,
Huang13}. Despite such a diversity, some methods are more frequently used~\cite{wu2008top}. Also, many of the commonly employed methods are defined in terms of similar assumptions about the data (e.g., k-means and k-medoids) or consider analogous mathematical concepts (e.g, similarity matrices for spectral or graph clustering) and, consequently, should provide similar performance in typical usage scenarios. Therefore, in the following we consider a choice of clustering algorithms from different families of methods. Several taxonomies have been proposed to organize the many different types of clustering algorithms into families~\cite{jain2004landscape, fraley1998many}. While some taxonomies categorize the algorithms based on their objective functions~\cite{jain2004landscape}, others aim at the specific structures desired for the obtained clusters (e.g. hierarchical)~\cite{fraley1998many}.  Here we consider the algorithms indicated in Table~\ref{tab:algorithm} as examples of the categories indicated in the same table. The algorithms represent some of the main types of methods in the literature. Note that some algorithms are from the same family, but in these cases they posses notable differences in their applications (e.g., treating very large datasets using clara). In Section~S1 of the supplementary material, we provide a short description about the parameters of each algorithm.

\begin{table}[h]
%d\center++++++++ing
\caption{\label{tab:algorithm}
{\bf Clustering methods considered in our analysis and the respective libraries and functions in R employing the methods.} The first column shows the name of the algorithms used throughout the text. The second column indicates the category of the algorithms. The third and fourth columns contain, respectively, the function name and R library of each algorithm.}
\small
\begin{tabular*}{\textwidth}{@{\extracolsep{\fill}}cccc}
	\toprule
	\multirow{2}{*}{ Algorithm name} & \multirow{2}{*}{Category} & \multirow{2}{*}{ Function in R} & \multirow{2}{*}{ Library in R}   \\ \\
	\hline
	k-means		 & Partitional		  & \textit{k-means} 				& stats\\
    clara   	 & Partitional		  & \textit{clara}	 				& cluster  \\
	hierarchical & Linkage			  & \textit{agnes}	 				& cluster\\
	EM 			 & Model-based		  & \textit{mstep}, \textit{estep} 	& mclust \\
	hcmodel 	 & Model-based		  & \textit{hc}		 				& mclust \\ 
	spectral 	 & Spectral methods					  & \textit{specc}	 				& kernlab  \\
	subspace 	 & Based on subspaces & \textit{hddc}	 				& HDclassif  \\
	\hline
	\hline
\end{tabular*}
\end{table}

\begin{figure*}[]
\centering
\includegraphics[width=\linewidth]{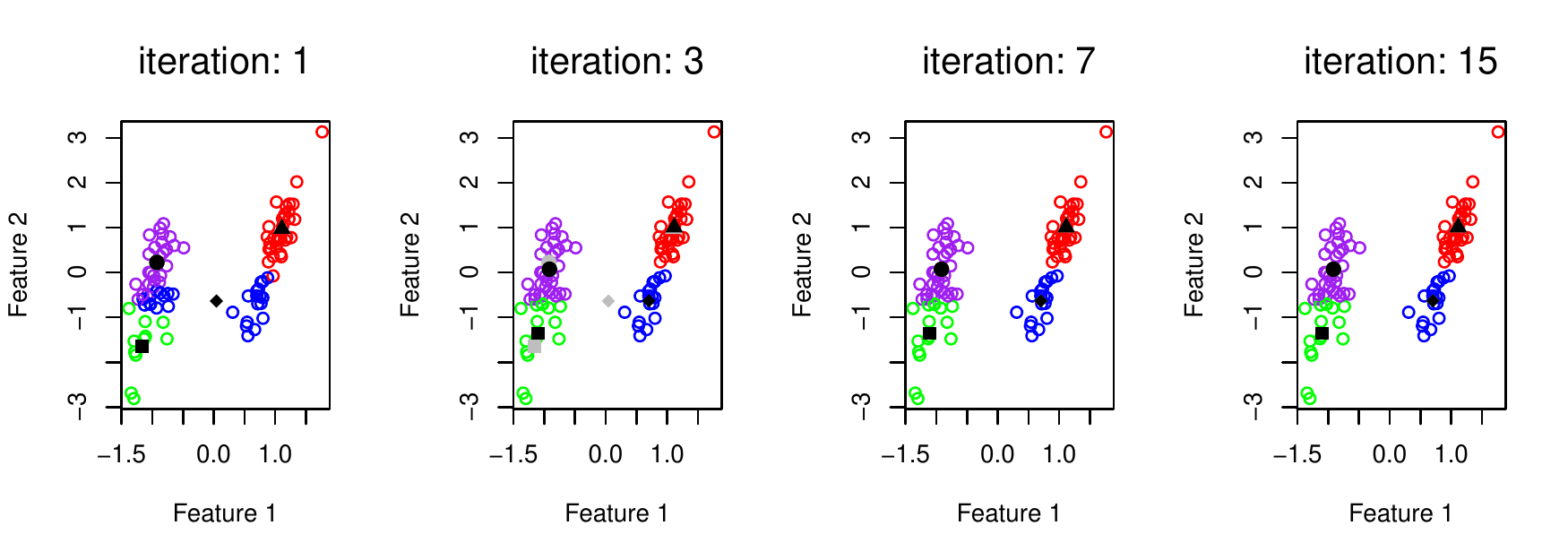} \\
\caption{\label{fig:Kmean}{\bf Illustration of the k-means clustering method.} Each plot shows the partition obtained after specific iterations of the algorithm. The centroids of the clusters are shown as a black marker. Points are colored according to their assigned clusters. Gray markers indicate the position of the centroids in the previous iteration. The dataset contains 2 clusters, but $k=4$ seeds were used in the algorithm.}  
\end{figure*}

Regarding partitional approaches, the k-means~\cite{macqueen1967some} algorithm is widely used by researchers~\cite{wu2008top}. This method requires as input parameters the number of groups ($k$) and a distance metric. Initially, each data point is associated with one of the $k$ clusters according to its distance to the centroids (clusters centers) of each cluster. An example is shown in Figure~\ref{fig:Kmean}(a), where black points correspond to centroids and the remaining points have the same color if the centroid that is closest to them is the same. Then, new centroids are calculated, and the classification of the data points is repeated for the new centroids, as indicated in Figure~\ref{fig:Kmean}(b), where gray points indicate the position of the centroids in the previous iteration. The process is repeated until no significant changes of the centroids positions is observed at each new step, as shown in Figures~\ref{fig:Kmean}(c) and (d). 

The \emph{a priori} setting of the number of clusters is the main limitation of the k-means algorithm. This is so because the final classification can strongly depend on the choice of the number of centroids~\cite{macqueen1967some}. In addition, the k-means is not particularly recommended in cases where the clusters do not show convex distribution or have very different sizes~\cite{Jain10, steinley2006k}.  Moreover, the k-means algorithm is sensitive to the initial  seed selection~\cite{jain1999data}. Given these limitations, many modifications of this algorithm have been proposed~\cite{dunn1973fuzzy, huangextensions,10.1371/journal.pone.0162259}, such as the k-medoid~\cite{kaufman2008clustering} and k-means$++$~\cite{arthur2007k}. Nevertheless, this algorithm, besides having low computational cost, can provide good results in many practical situations such as in anomaly detection~\cite{sequeira2002admit} and data segmentation~\cite{williams1997mining}. The R routine used for k-means clustering was the \textit{k-means} from the \textit{stats} package.

Another interesting example of partitional clustering algorithms is the clustering for large applications (clara) \cite{kaufman2008clustering}. This method takes into account multiple fixed samples of the dataset to minimize sampling bias and, subsequently, select the best medoids among the chosen samples, where a medoid is defined as the object $i$ for which the average dissimilarity to all other objects in its cluster is minimal. This method is efficient for large amounts of data because it does not explore the whole neighborhood of the data points~\cite{Han2006}, although the quality of the results have been found to strongly depend on the number of objects in the sample data~\cite{huangextensions}. The clara algorithm employed in our analysis was from the \textit{clara} function contained in the \textit{cluster} package.

Clustering methods that take into account the linkage between data points, traditionally known as hierarchical methods, can be subdivided into two groups: agglomerative and divisive~\cite{Jain10}. In an agglomerative hierarchical clustering algorithm, initially, each object belongs to a respective individual cluster. Then, after successive iterations, groups are merged until stop conditions are reached. On the other hand, a divisive hierarchical clustering method starts with all objects in a single cluster and, after successive iterations, objects are separated into clusters. There are two main packages in the R language that provide routines for performing hierarchical clustering, they are the \emph{stats} and \emph{cluster}. Here we consider the \textit{agnes} routine from the \emph{cluster} package. Four well-known linkage criteria are available in \textit{agnes}, namely single linkage, complete linkage, Ward’s method, and weighted average linkage~\cite{lance1967general}.
  
Model-based methods can be regarded as a general framework for estimating the maximum likelihood of the parameters of an underlying distribution to a given dataset. A well-known instance of model-based methods is the expectation-maximization algorithm (EM). Most commonly, one considers that the data from each class can be modeled by multivariate normal distributions, and, therefore, the distribution observed for the whole data can be seen as a mixture of such normal distributions. A maximum likelihood approach is then applied for finding the most probable parameters of the normal distributions of each class. The EM approach for clustering is particularly suitable when the dataset is incomplete~\cite{Redner84,Dempster77}. On the other hand, the clusters obtained from the method may strongly depend on the initial conditions~\cite{aggarwal2013data}. In addition, the algorithm may fail to find very small  clusters~\cite{fraley1998many, fraley2002model}. In the R language, functions \textit{estep} and \textit{mstep} from the \textit{mclust}~\cite{fraley1999mclust,Fraley2003} package can be used for employing the EM method. A related algorithm that is also analyzed in the current study is the hcmodel, which can be found in the \textit{hc} function of the \textit{mclust} package. The hcmodel algorithm is also based on Gaussian-mixture evaluation, but it contains many additional steps such as an agglomerative procedure and the adjustment of model parameters through a Bayes factor selection with the BIC aproximation~\cite{schwarz1978estimating}. 

Another class of methods considered in our analyses is spectral clustering. These methods emerged as an alternative to traditional clustering approaches that were not able to define nonlinear discriminative hypersurfaces~\cite{nascimento2011spectral}. The main advantage of spectral methods lies on the definition of an adjacency structure from the original dataset, which avoids imposing a prefixed shape for the clusters~\cite{filippone2008survey}. The first step of the method is to construct an affinity  matrix $A \in \mathbb{R}^{N x N}$, where the value in the $j$-th row and $k$-th column indicates the similarity between points $j$ and $k$. This matrix can be regarded as a weighted graph representation of the data. Then, the eigenvalues and eigenvectors of the matrix are used for partitioning the data according to a given criterion. Many different types of similarity matrices can be used, a popular choice being the Laplacian matrix~\cite{von2007tutorial}. One disadvantage of spectral methods is the costly process of calculating the eigenvectors of the similarity matrix~\cite{aggarwal2013data}. The function \emph{specc} from the \emph{kernlab} R package, which employs a kernel function to compute the affinity matrix, was used for evaluating the performance of the spectral method.

In recent years, the efficient handling of high dimensional data has become of paramount importance and, for this reason, this feature has been desired when choosing the most appropriate method for obtaining accurate partitions. To tackle high dimensional data, subspace clustering was proposed~\cite{parsons2004subspace}. This method works by considering the similarity between objects with respect to distinct subsets of the attributes~\cite{kriegel2012subspace}. The motivation for doing so is that different subsets of the attributes might define distinct separations between the data. Therefore, the algorithm can identify clusters that exist in multiple, possibly overlapping, subspaces~\cite{parsons2004subspace}. Subspace algorithms can be categorized into four main families~\cite{sim2013survey}, namely: lattice, statistical, approximation and hybrid. The \textit{hddc} function from package \textit{HDclassif} implements the subspace clustering method in the R language. The algorithm is based on statistical models, with the assumption that all attributes may be relevant for clustering~\cite{bouveyron2007high}. Some parameters of the algorithm, such as the number of clusters or model to be used, are estimated using an EM procedure.

\section{Materials and Methods}

\subsection{Artificial datasets}
\label{s:artfData}

The proper comparison of clustering algorithms requires a robust artificial data generation method to produce a variety of datasets. For such a task, we apply a  methodology based on a previous work by Hirschberger et al.~\cite{hirschberger2007randomly}. The procedure can be used to generate samples characterized by $F$ features and separated into $C$ classes. In addition, the method can control both the  variance and correlation distributions among the features for each class. The artificial dataset can also be generated  by varying the number of objects per class, $N_e$, and the expected separation, $\alpha$, between the classes. 

The main difficult in generating datasets with the aforementioned properties is the definition of a proper covariance matrix $R$ for the considered features. A valid covariance matrix must be positive semi-definite~\cite{horn2012matrix}, which is hard to ensure. However, for a given matrix $\mathbf{G}\in \mathbb{R}^{n\times m}$, the matrix $R=\mathbf{G}\mathbf{G}^T$ is guaranteed to be positive semi-definite~\cite{horn2012matrix}. Thus any random matrix $\mathbf{G}$ can define a valid respective covariance matrix. As a consequence, additional constraints on matrix $\mathbf{G}$ can be imposed for the generation of datasets with the required properties. Hirschberger et al.~\cite{hirschberger2007randomly} developed a robust approach to generate such a matrix given the first two statistical moments of the co-variance distribution of a set of $F$ artificial features. The resulting covariance matrix contains variances and co-variances drawn from such distribution. Here we consider a normal distribution to represent the elements of $\mathbf{R}$. 

For each class $i$ in the dataset, a covariance matrix $\mathbf{R}_i$ of size $F\times F$ is generated, and this matrix is used to define $N_e$ objects for the classes. This means that pairs of features can have distinct correlation for each generated class. Then, the generated class values are divided by $\alpha$ and translated by $s_i$, where $s_i$ is a random variable described by a uniform random distribution defined in the interval $[-1,1]$. Parameter $\alpha$ is associated with the expected distances between classes. Such distances can have different impacts on clusterization depending on the number of objects and features used in the dataset. Notice that such a procedure for the generation of artificial datasets was previously used in~\cite{amancio2014systematic}.

\begin{figure*}[h]
\centering
\includegraphics[width=0.9\linewidth]{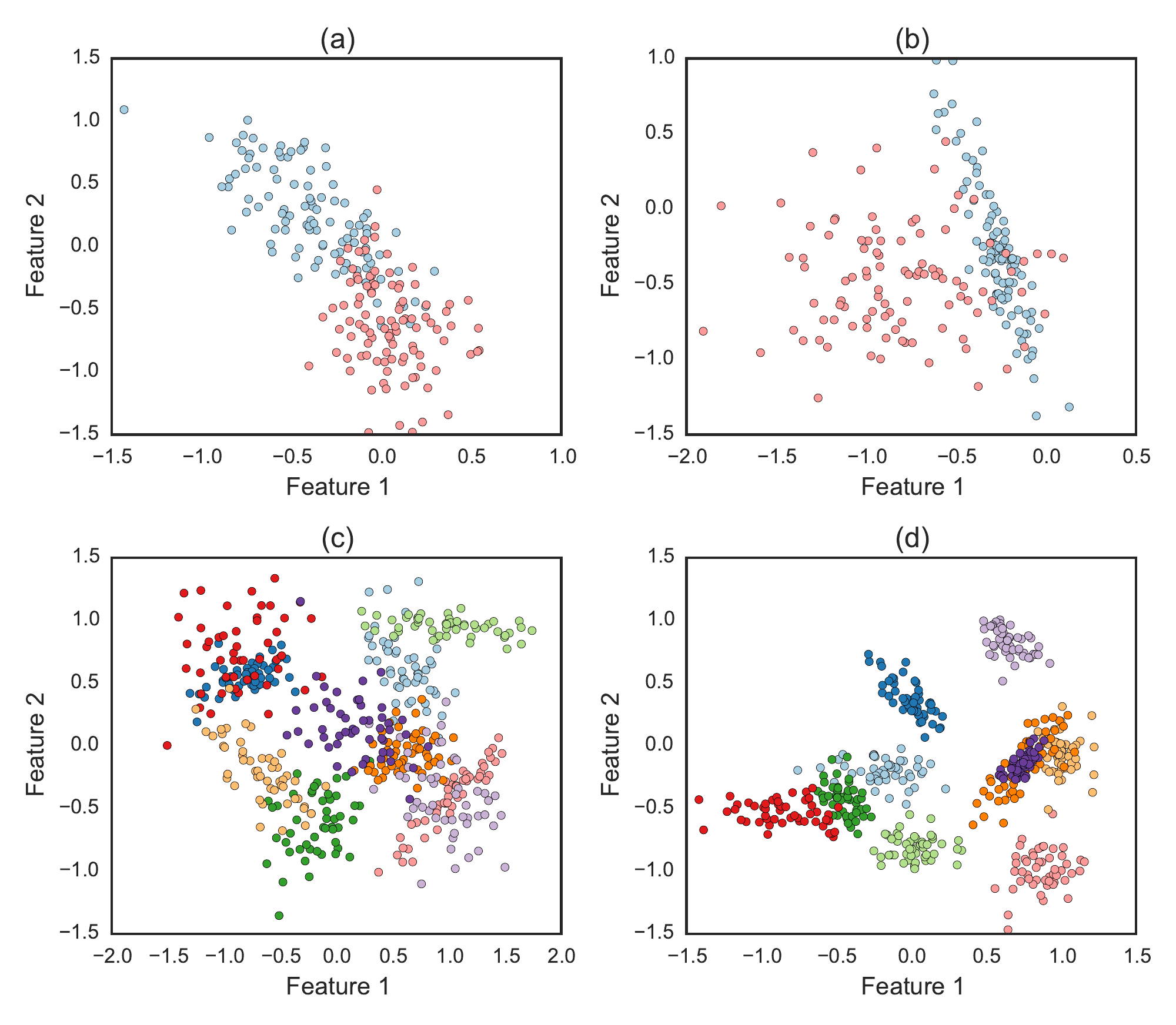}
\caption{{\bf Examples of artificial datasets generated by the methodology.} The parameters used for each case are (a) $C$=2, $N_e$=100 and $\alpha$=3.3. (b) $C$=2, $N_e$=100 and $\alpha$=2.3. (c) $C$=10, $N_e$=50 and $\alpha$=4.3. (d) $C$=10, $N_e$=50 and $\alpha$=6.3. Note that each class can present highly distinct properties due to differences in correlation between their features.} 
\label{fig:dataExamples}
\end{figure*}

In Figure~\ref{fig:dataExamples}, we show some examples of artificially generated data. For visualization purposes, all considered cases contain $F=2$ features. The parameters used for each case are described in the caption of the figure. Note that the methodology can generate a variety of dataset configurations, including variations in features correlation for each class.

In this study, we considered the following values for the artificial dataset parameters:
\begin{itemize}
 
 \item {\bf Number of classes} ($C$): The generated datasets are divided into  $C=\{2,10,50\}$ classes. 
 
 \item {\bf Number of features} ($F$): The number of features to characterize the objects is $F=\{2,10,50\}$. 
 
 \item {\bf Number of object per class} ($N_e$): we considered $Ne=\{5,50,100\}$ objects per class. In our experiments, in a given generated dataset, the number of instances for each class is constant.
 
\item {\bf Mixing parameter} ($\alpha$): This parameter has a non-trivial dependence on the number of classes and features. Therefore, for each dataset, the value of this parameter was tuned so that no algorithm would achieve an accuracy of 0\% or 100\%.

\end{itemize}

We refer to datasets containing 2, 10 and 50 features as DB2F, DB10F, DB50F, respectively. Such datasets are composed of all considered number of classes, C=$\{2,10,50\}$, and 50 elements for each class (i.e., Ne=50). In some cases, we also indicate the number of classes considered for the dataset. For example, dataset DB2C10F contains 2 classes, 10 features and 50 elements per class. For each case, we consider 10 realizations of the dataset. Therefore, 270 datasets were generated in total.

\subsection{Evaluating the performance of clustering algorithms}

The evaluation of the quality of the generated partitions is one of the most important issues in cluster analysis~\cite{halkidi2001clustering}. 
Here, we adopt the most traditional indexes, which are the Jaccard Index (J)~\cite{citeulike:7707101}, Adjusted Rand Index (ARI)~\cite{Hubert1985}, Fowlkes Mallows Index (FM)~\cite{Fowlkes83} and Normalized Mutual Information (NMI)~\cite{Strehl02clusterensembles}.

In order to define the cluster quality metrics, we consider the following concepts. Let $U = \left\lbrace u_1, u_2 ... u_R \right\rbrace $ represent the original partition of a dataset, where $u_i$ denote a subset of the objects associated with cluster $i$. Equivalently, let $V = \left\lbrace v_1, v_2 ... v_C \right\rbrace $ 
represent the partition found by a cluster algorithm. We denote as $a$ the number of pairs of objects that are placed in the same group in both $U$ and $V$. Mathematically, $a$ can be computed by 
\begin{equation}
	a=\sum_{i,j} \binom{n_{ij}}{2},
\end{equation}
where $n_{ij}$ is the number of objects belonging to both subset $u_i$ and $v_j$. \\
Let $b$ indicate the number of pairs of objects belonging to the same group in $U$ but different groups in $V$, i.e. 
\begin{equation}
	b=\sum_{i} \binom{n_{i.}}{2}- \sum_{i,j} \binom{n_{ij}}{2},
\end{equation}
where $n_{i.}=\sum_{j}n_{ij}$. 
Let $c$ be the number of pairs of objects belonging to different groups in $U$ and to the same group in $V$, which can be written as 
\begin{equation}
	c=\sum_{j} \binom{n_{.j}}{2}- \sum_{i,j} \binom{n_{ij}}{2},
\end{equation}
where $n_{.j}=\sum_{i}n_{ij}$.

The Jaccard Index ($J$), Adjusted Rand Index (ARI) and Fowlkes Mallows (FM) index  can then be defined based on $a$, $b$, $c$: %according to the equations presented in Table~\ref{tab:perfMeaDef}.
\begin{equation}
	J = \frac{a}{ a+b+c},
\end{equation}
\begin{equation}
	\textrm{ARI} = \frac{\sum_{i,j} \binom{n_{ij}}{2} - \left[  \sum_{i} \binom{n_{i.}}{2} \sum_{j} \binom{n_{.j}}{2} \right] / \binom{n}{2} } {1/2 \left[  \sum_{i} \binom{n_{i.}}{2} + \sum_{j} \binom{n_{.j}}{2} \right] - \left[  \sum_{i} \binom{n_{i.}}{2} \sum_{j} \binom{n_{.j}}{2} \right] / \binom{n}{2}},
\end{equation}
\begin{equation}
	\textrm{FM}= a {\frac{\sqrt{a+b}\sqrt{a+c}}{(a+b)(a+c)}  }.
\end{equation}
We also consider the normalized mutual information(NMI) as a quality metric because it
quantifies the mutual dependence between two random
variables based on well-established concepts of information theory~\cite{cover2012elements}. The NMI measure is defined as~\cite{strehl2002cluster}
\begin{equation}
	\textrm{NMI}(C,T) = \frac{I(C, T)}{\sqrt{[H(C), H(T)]}}.
\end{equation}
where $C$ is the random variable denoting the
cluster assignments of the points, and $T$ is the
random variable denoting the underlying class
labels on the points.
$I(C,T)=  H(C) - H(C|T)$  is the mutual information between the random variables $C$ and $T$. $H(C)$ is the Shannon entropy of $C$. $H(C|T)$ is the conditional entropy of $C$ given $T$. 

Note that when the two sets of labels have a perfect one-to-one correspondence, the quality measures are all equal to unity. 
  
\section{Results and Discussion}

The accuracy of each considered clustering algorithm was evaluated using three methodologies. In the first methodology, we consider the default parameters of the algorithms provided by the R package. The reason for measuring performance using the default parameters is to consider the case where a researcher  applies the classifier to a dataset without any parameter adjustment. This is a common scenario when the researcher is not a machine learning expert. %
In the second methodology, we quantify the influence of the algorithms parameters on the accuracy. This is done by varying a single parameter of an algorithm while keeping the others at their default values. The third methodology consists in analyzing the performance by randomly varying all parameters of a classifier. This procedure allows the quantification of certain properties such as the maximum accuracy attained and the sensibility of the algorithm to parameter variation. 

\subsection{Performance when using default parameters}

In this experiment, we evaluated the performance of the classifiers for all datasets described in Section~\ref{s:artfData}. All unsupervised algorithms were set with their default configuration of parameters. For each algorithm, we divide the results according to the number of features contained in the dataset. In other words, for a given number of features, $F$, we used datasets with $C=\{2,10,50\}$ classes, and $N_e=\{5,50,100\}$ objects for each class. Thus, the performance results obtained for each $F$ corresponds to the performance averaged over distinct number of classes and objects per class.  
We note that the algorithm based on subspaces cannot be applied to datasets containing 2 features, and therefore its accuracy was not quantified for such datasets. 

\begin{figure*}
    \begin{center}
          \begin{subfigure}[][]{
	      {\includegraphics[width=0.45\columnwidth]{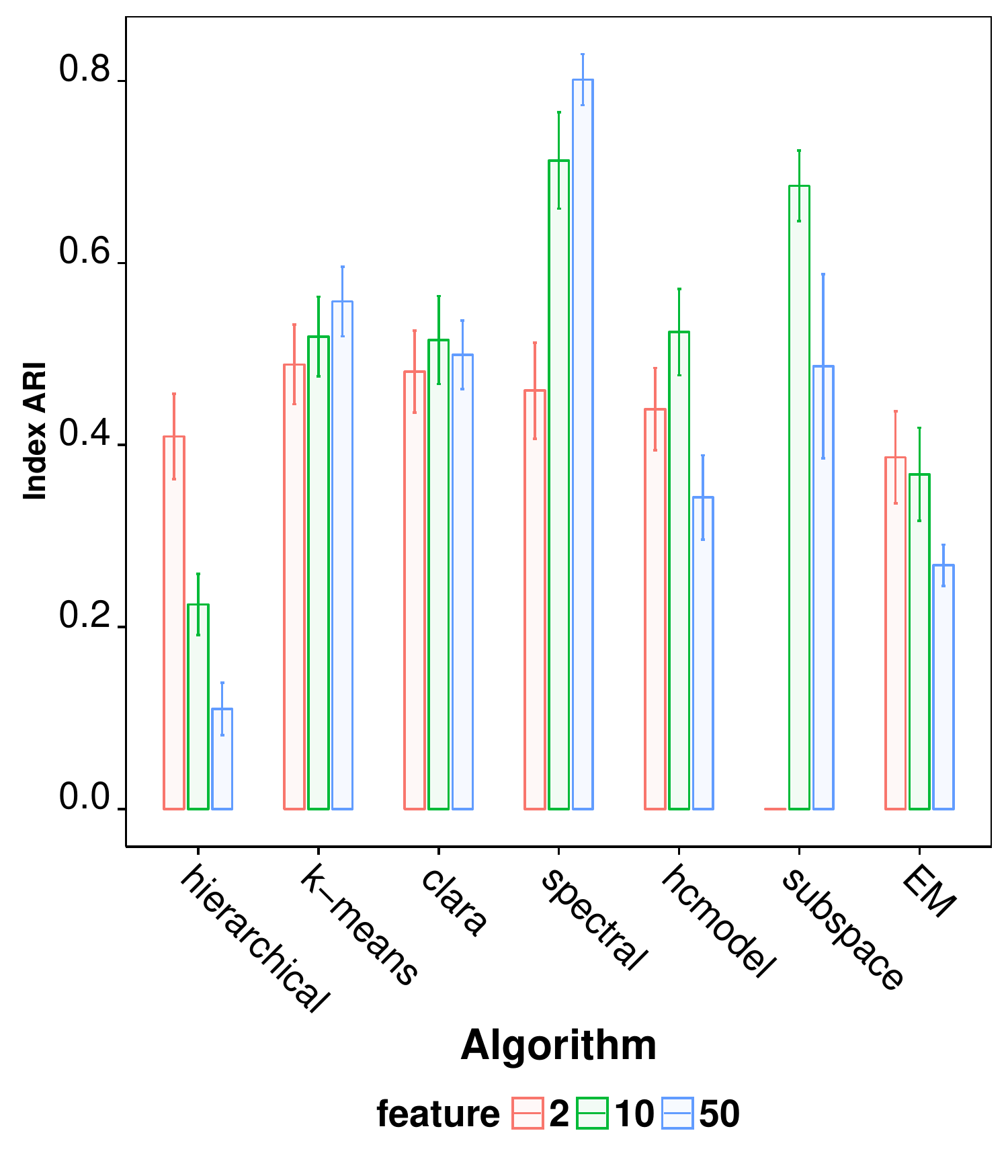}}\label{fig:IndexARI_Feature}}
          \end{subfigure}
          \begin{subfigure}[][]{
	      {\includegraphics[width=0.45\columnwidth]{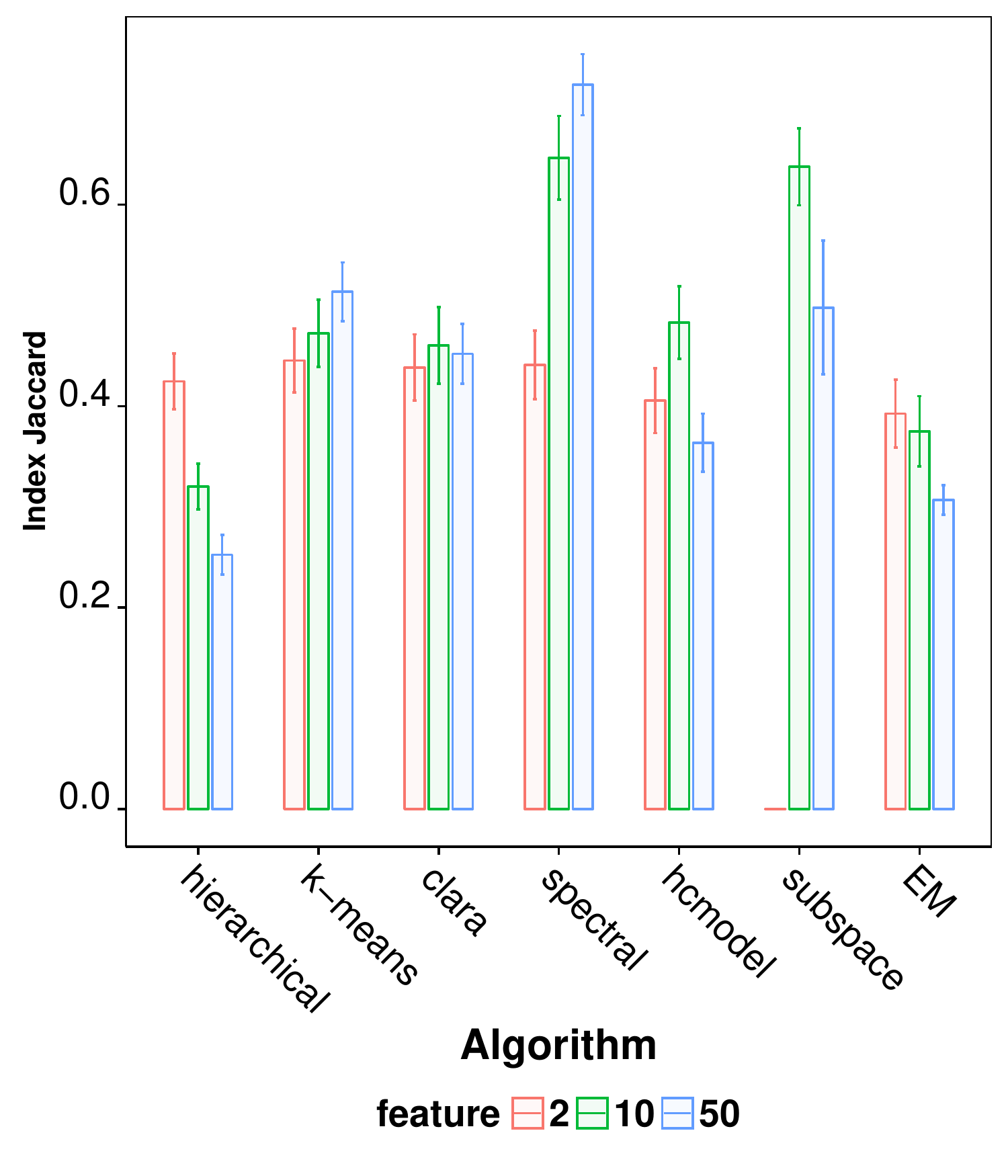}}\label{fig:IndexJaccard_Feature}}
          \end{subfigure}
          \begin{subfigure}[][]{
	      {\includegraphics[width=0.45\columnwidth]{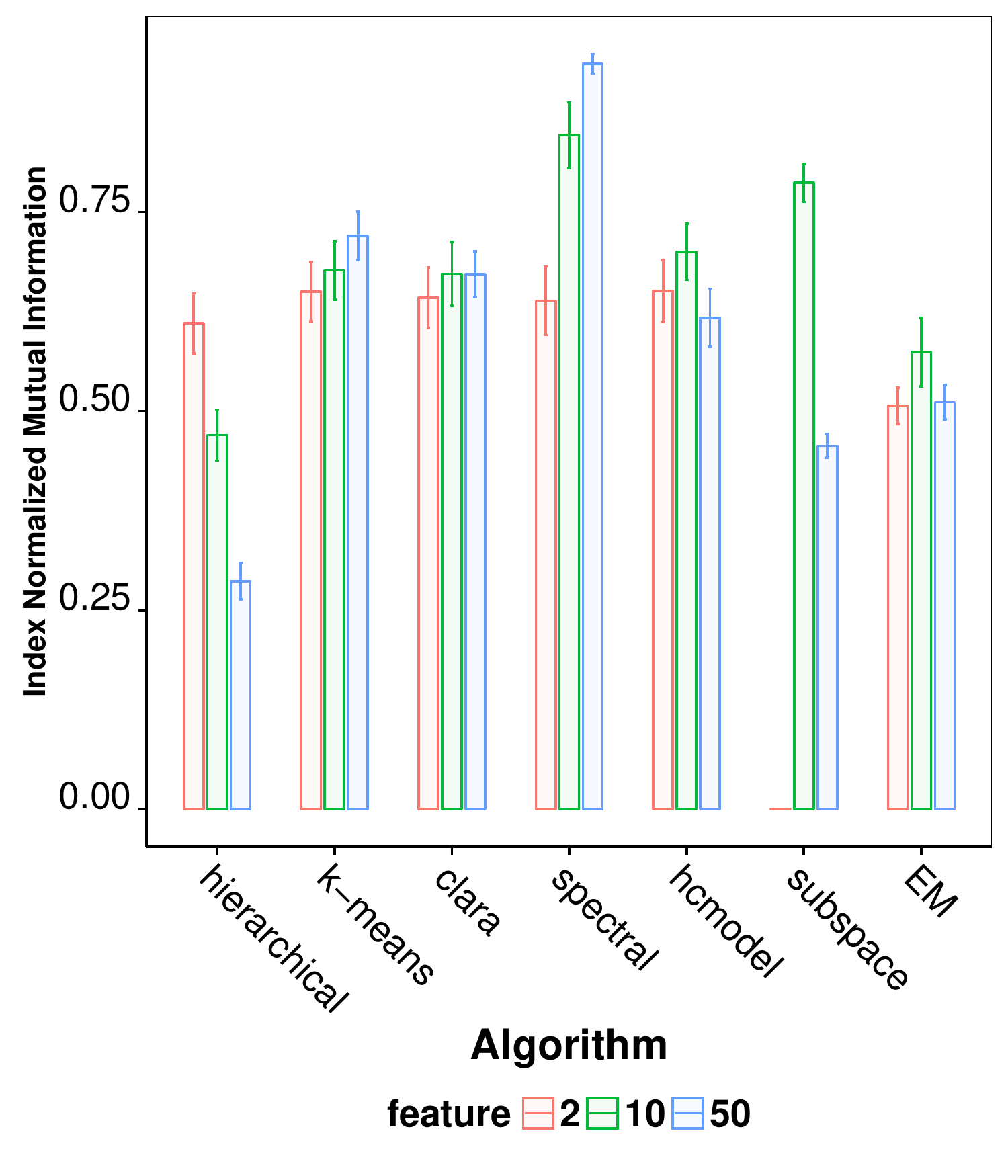}}\label{fig:NMI_Feature}}
          \end{subfigure}
           \begin{subfigure}[][]{
	      {\includegraphics[width=0.45\columnwidth]{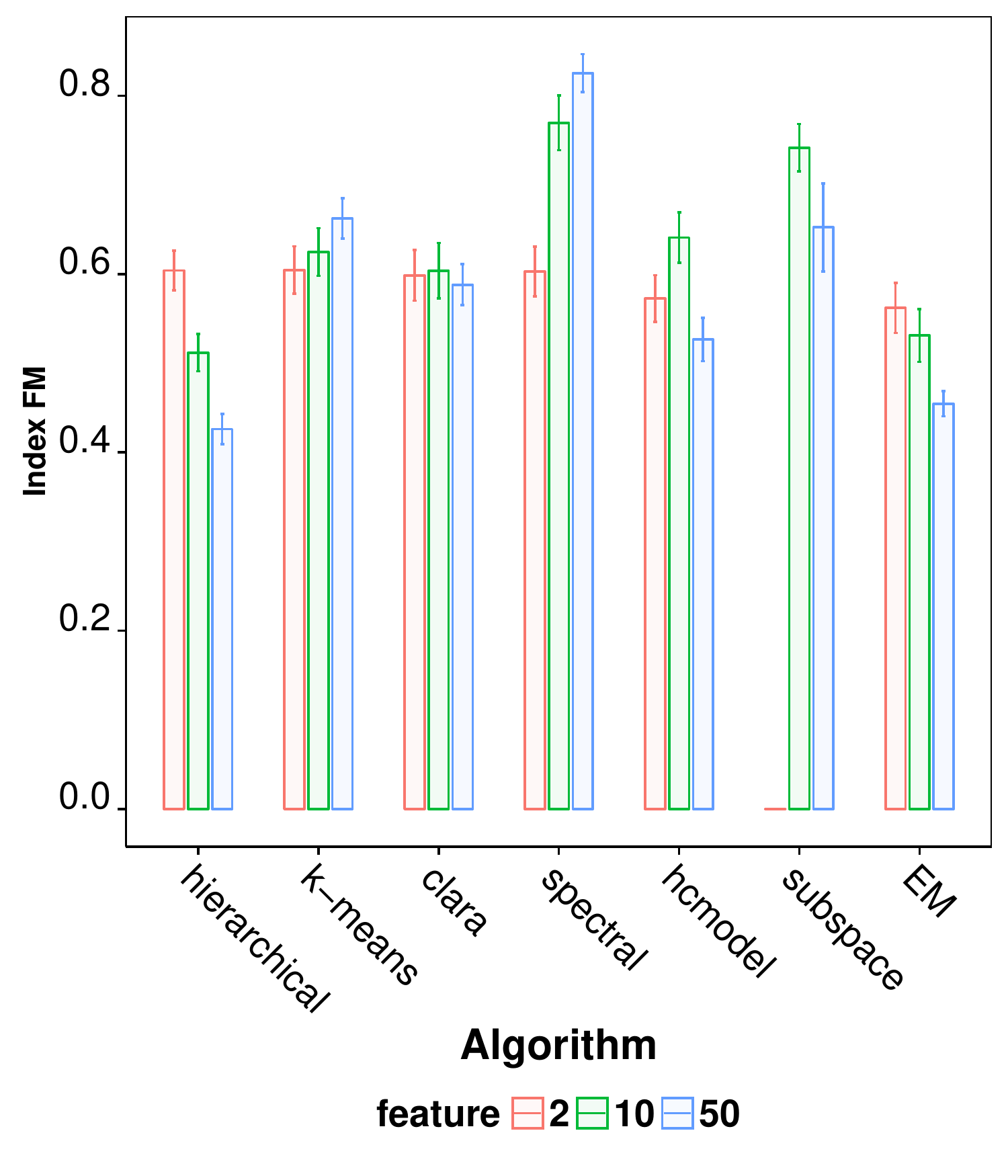}}\label{fig:IndexFM_Feature}}
          \end{subfigure}
         
   \end{center}
\caption{{\bf Average performance of the seven considered clustering algorithms according to the number of features in the dataset.} All datasets described in Section~\ref{s:artfData} were used for evaluation. The averages were calculated separately for datasets containing 2, 10 and 50 features. The considered performance indexes are (a) adjusted Rand, (b) Jaccard, (c) normalized mutual information and (d) Fowlkes Mallows.}
\label{fig:AllDatabase_Feature}
\end{figure*}

In Figure~\ref{fig:AllDatabase_Feature}, we show the obtained values for the four considered performance metrics.
The results indicate that all performance metrics provide similar results. Also, the hierarchical method seems to be strongly affected by the number of features in the dataset. In fact, when using 50 features the hierarchical method provided the worst results among all methods. The k-means and spectral methods benefit from an increment in the number of features. Interestingly, the hcmodel has a better performance in the datasets containing 10 features than in those containing 2 and 50 features, which suggests an optimum performance for this algorithm for datasets containing around 10 features.
It is also clear that for 2 features all algorithms display similar performance, while a larger number of features induce marked differences in performance. In particular, for 50 features, the spectral algorithm provides the best results among all classifiers.

We use the Kruskal-Wallis test~\cite{mckight2010kruskal}, a one-way ANOVA nonparametric test, to explore the statistical differences in performance when considering distinct number of features in clustering methods. First, we test if the difference in performance is significant for 2 features. For this case, the Kruskal-Wallis test returns a p-value of $p = 0.07$, with a chi-squared distance of $\chi^2= 10.26$. Therefore, the difference in performance do not seem to be statistically significant. When considering the results for 10 features, a p-value of $p = 4.4\times 10^{-6}$ is returned by the Kruskal-Wallis test, with a chi-squared distance of $\chi^2= 34.94$). For 50 features, the test returns a p-value of $p= 1.4\times 10^{-6}$, with a chi-squared distance of $\chi^2= 37.48$). This means that, in contrast for the case of 2 features, the algorithms indeed have significant differences in performance for 10 and 50 features, as indicated in Figure~\ref{fig:AllDatabase_Feature}.

\begin{figure*}[]
    \begin{center}
          
          \begin{subfigure}[][]{
	      {\includegraphics[width=0.45\columnwidth]{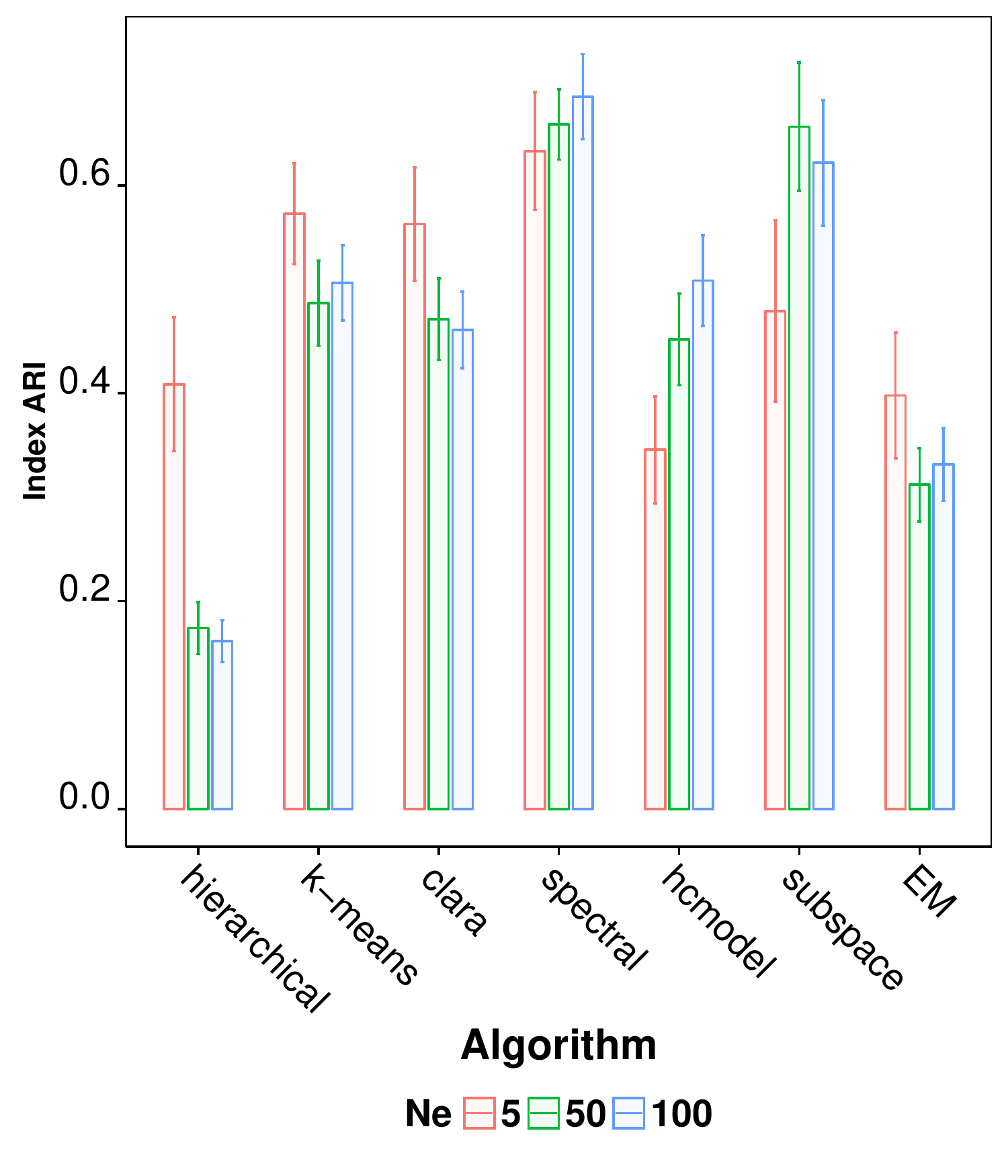}}\label{fig:IndexARI_Ne}}
          \end{subfigure}
          \begin{subfigure}[][]{
	      {\includegraphics[width=0.45\columnwidth]{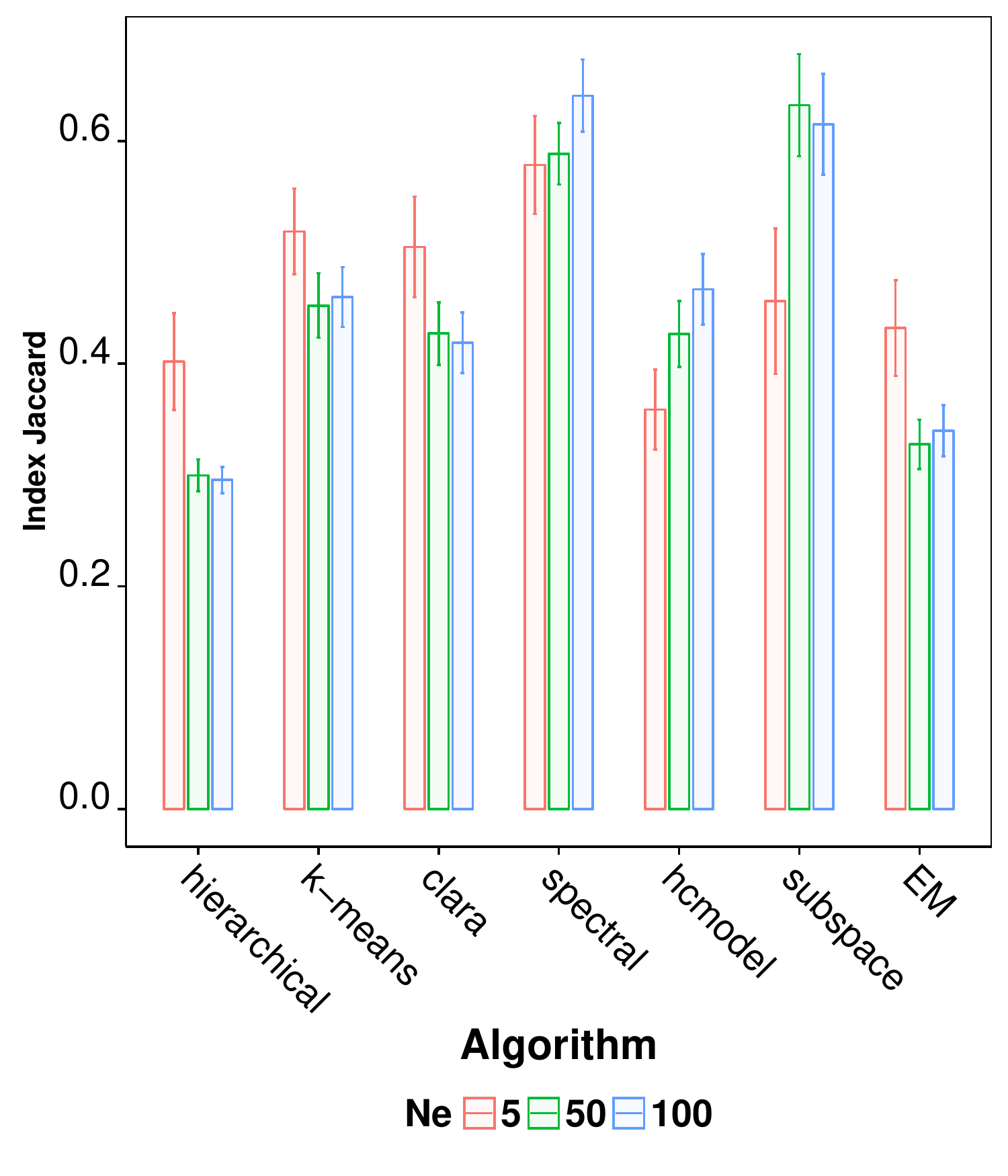}}\label{fig:IndexJaccard_Ne}}
          \end{subfigure}
          \begin{subfigure}[][]{
	      {\includegraphics[width=0.45\columnwidth]{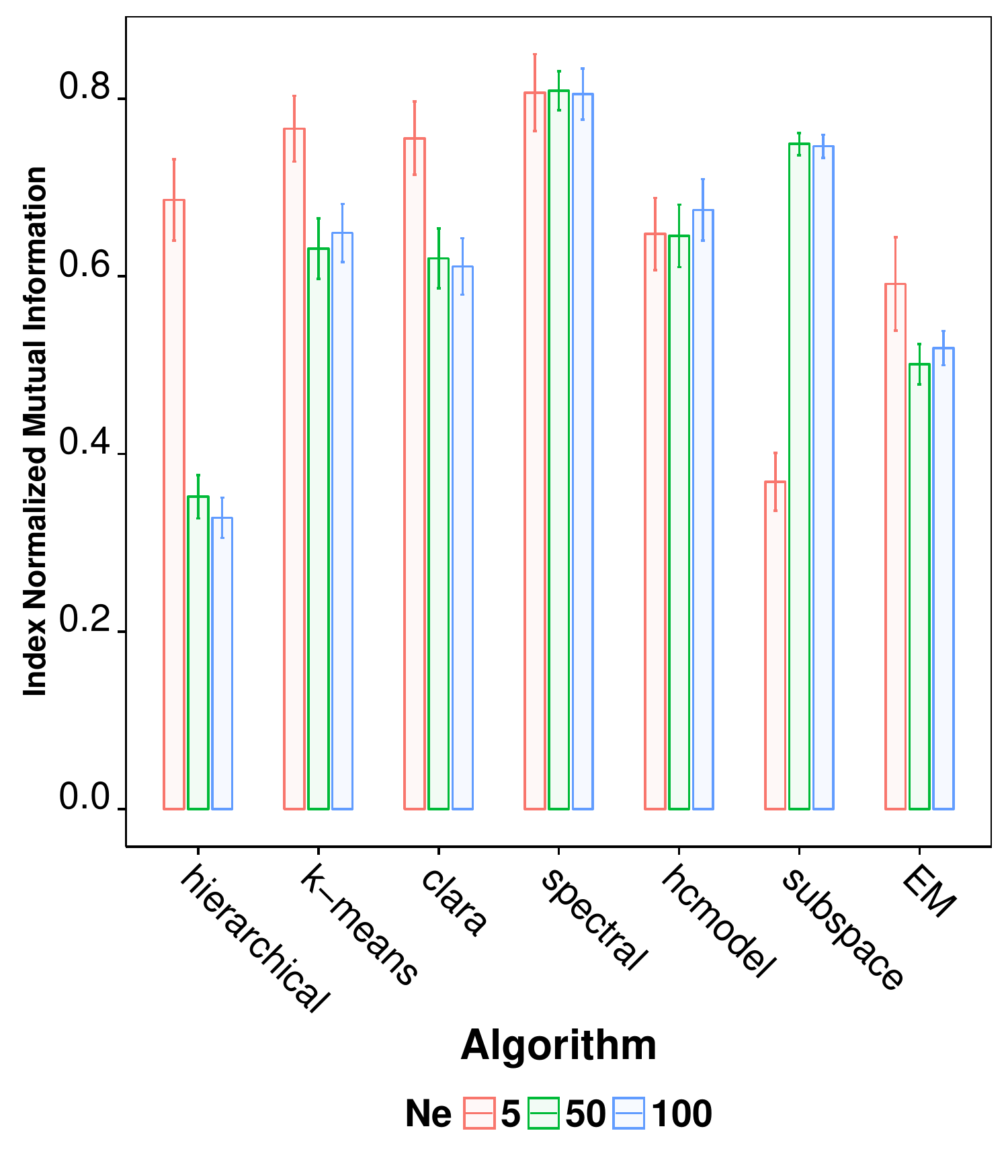}}\label{fig:NMI_Ne}}
          \end{subfigure}
           \begin{subfigure}[][]{
	      {\includegraphics[width=0.45\columnwidth]{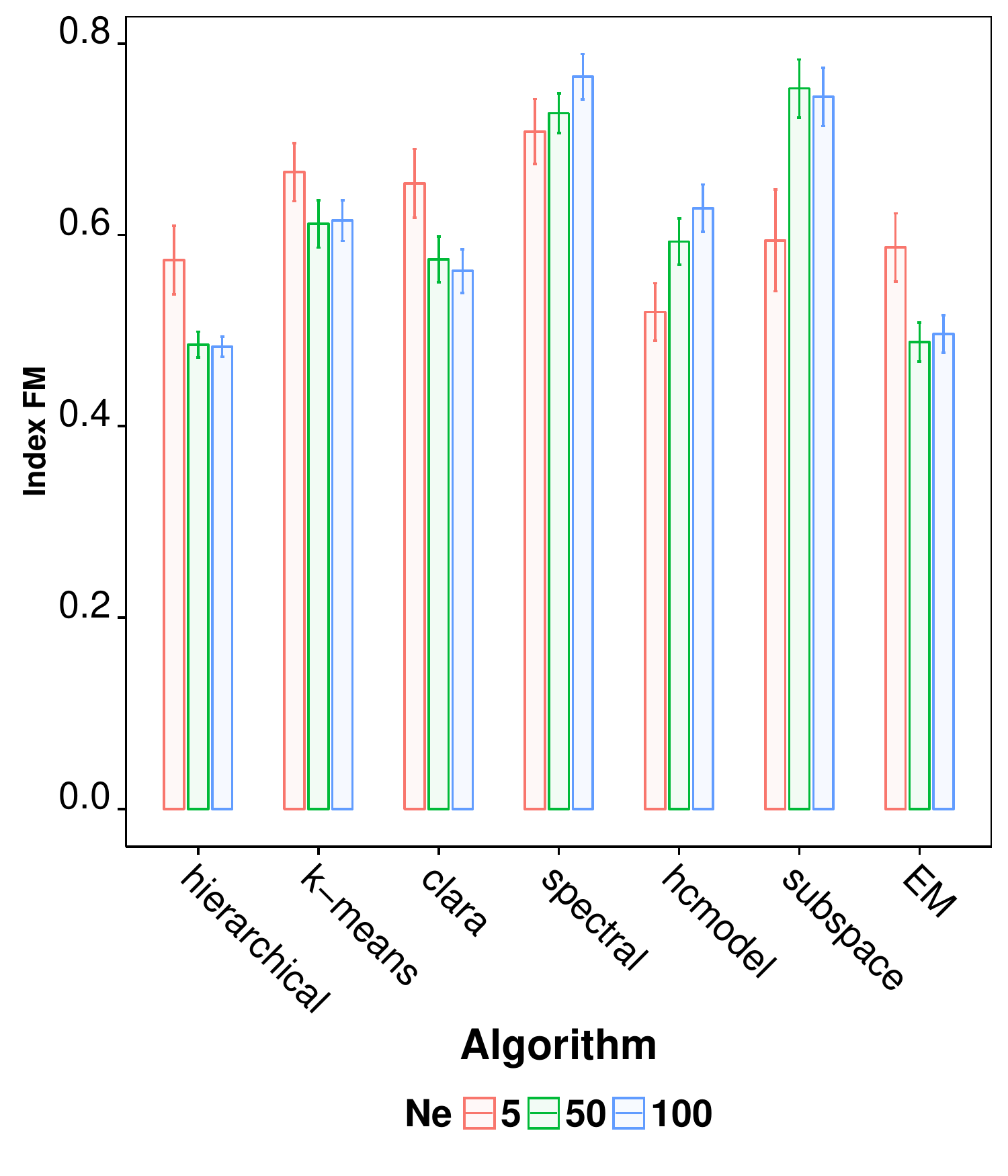}}\label{fig:IndexFM_Ne}}
          \end{subfigure}
   \end{center}
\caption{{\bf Average performance of the seven considered clustering algorithms according to the number of objects per class in the dataset.} All datasets described in Section~\ref{s:artfData} were used for evaluation. The averages were calculated separately for datasets containing 5, 50 and 100 objects per class. The considered performance indexes are (a) adjusted Rand, (b) Jaccard, (c) normalized mutual information and (d) Fowlkes Mallows.}
\label{fig:AllDatabase_Ne}
\end{figure*}

In order to verify the influence of the number of objects used for classification, we also calculated the average accuracy for datasets separated according to the number of objects $N_e$. The result is shown in Figure~\ref{fig:AllDatabase_Ne}. We observe that the impact that changing $N_e$ has on the accuracy depends on the algorithm. Surprisingly, the hierarchical, k-means and clara methods attain lower accuracy when more data is used. The result indicates that these algorithms are less robust with respect to the larger overlap between the clusters due to an increase in the number of objects. We also observe that a larger $N_e$ markedly benefits the performance of the subspace method. This results is in agreement with~\cite{hal00541203}.

\begin{figure*}
    \begin{center}
          \begin{subfigure}[][]{
	      {\includegraphics[width=0.45\columnwidth]{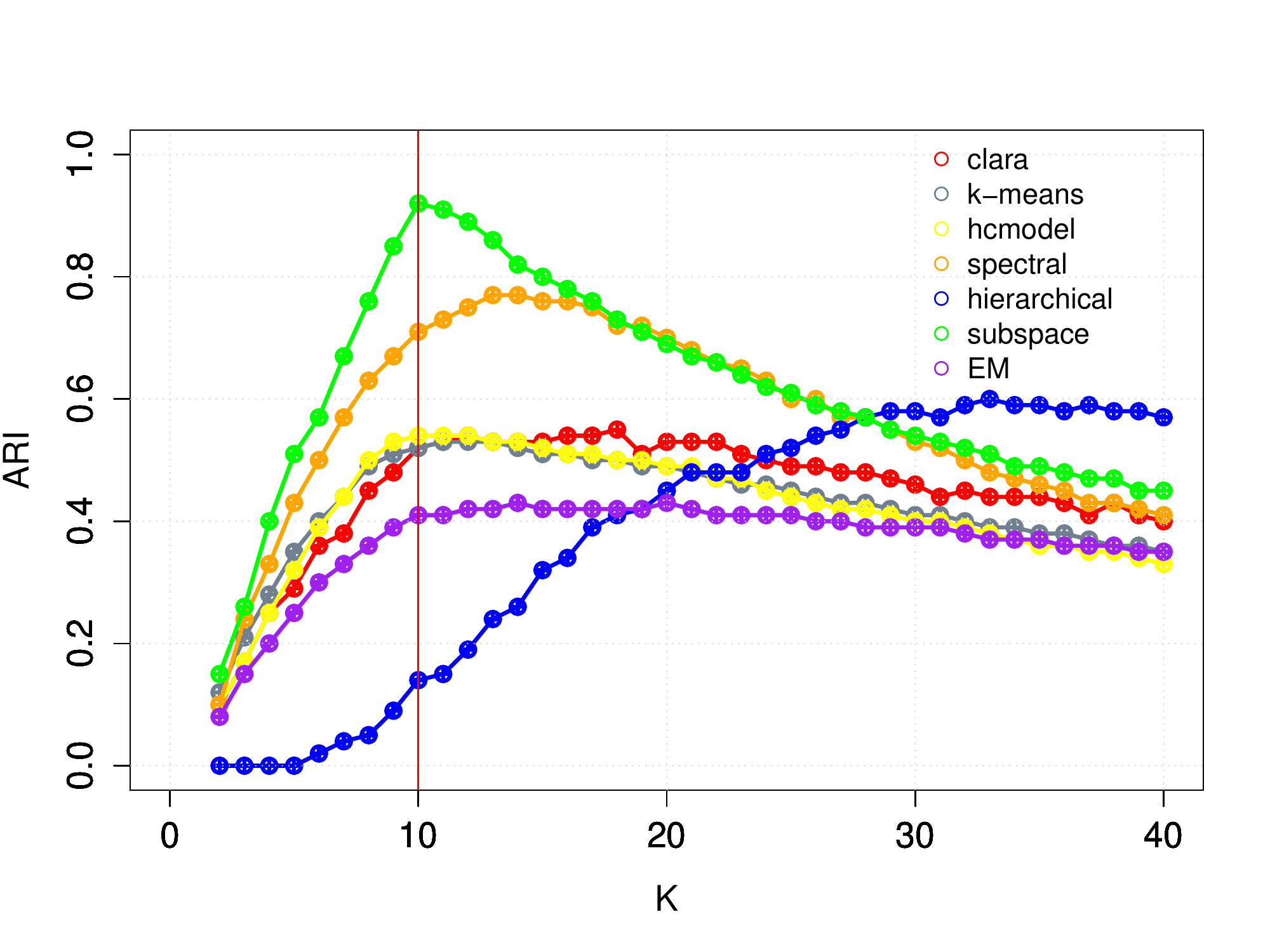}}\label{fig:clusteringKa}}
          \end{subfigure}
          \begin{subfigure}[][]{
	      {\includegraphics[width=0.45\columnwidth]{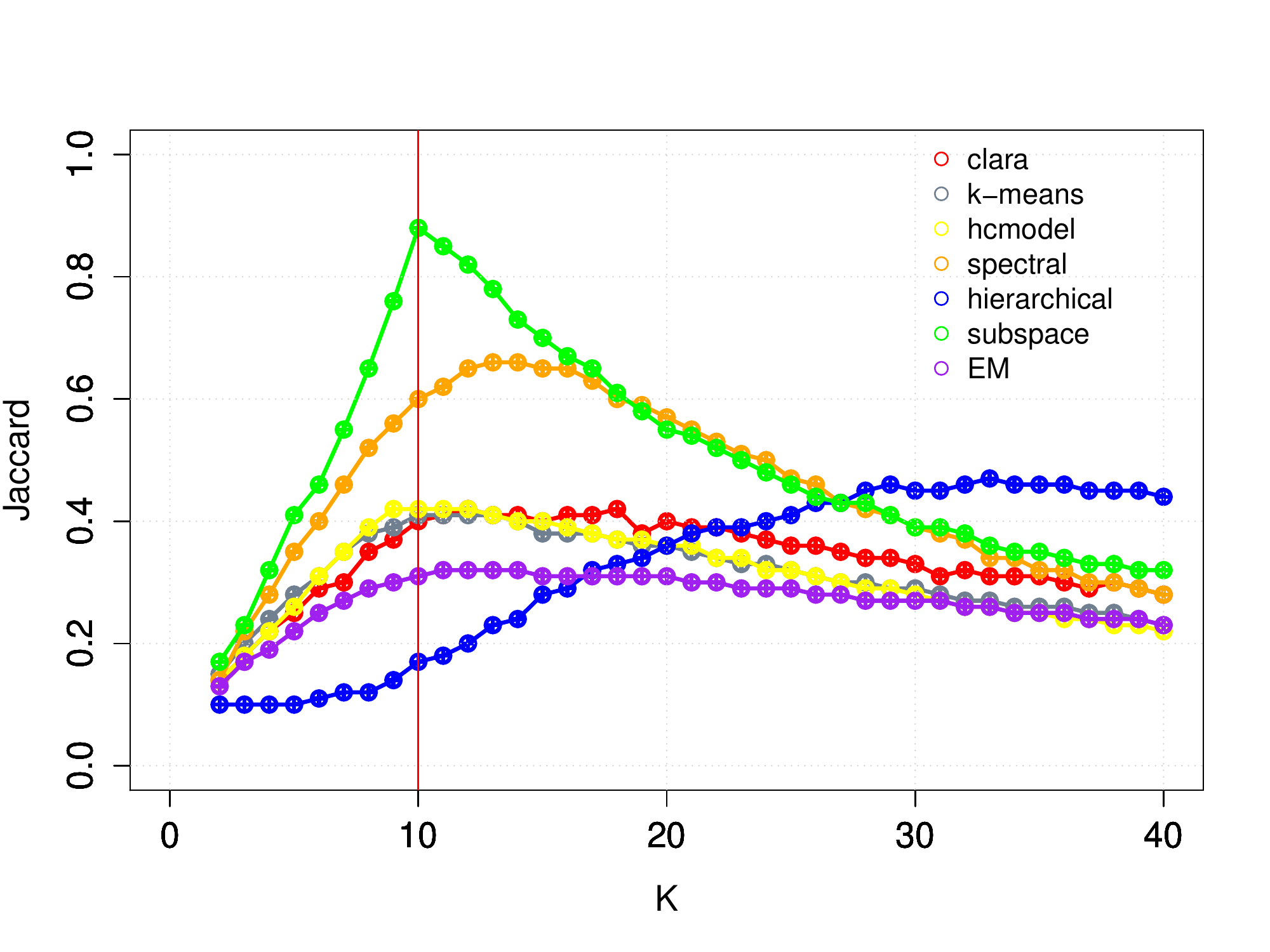}}\label{fig:clusteringKb}}
          \end{subfigure}
          \begin{subfigure}[][]{
	      {\includegraphics[width=0.45\columnwidth]{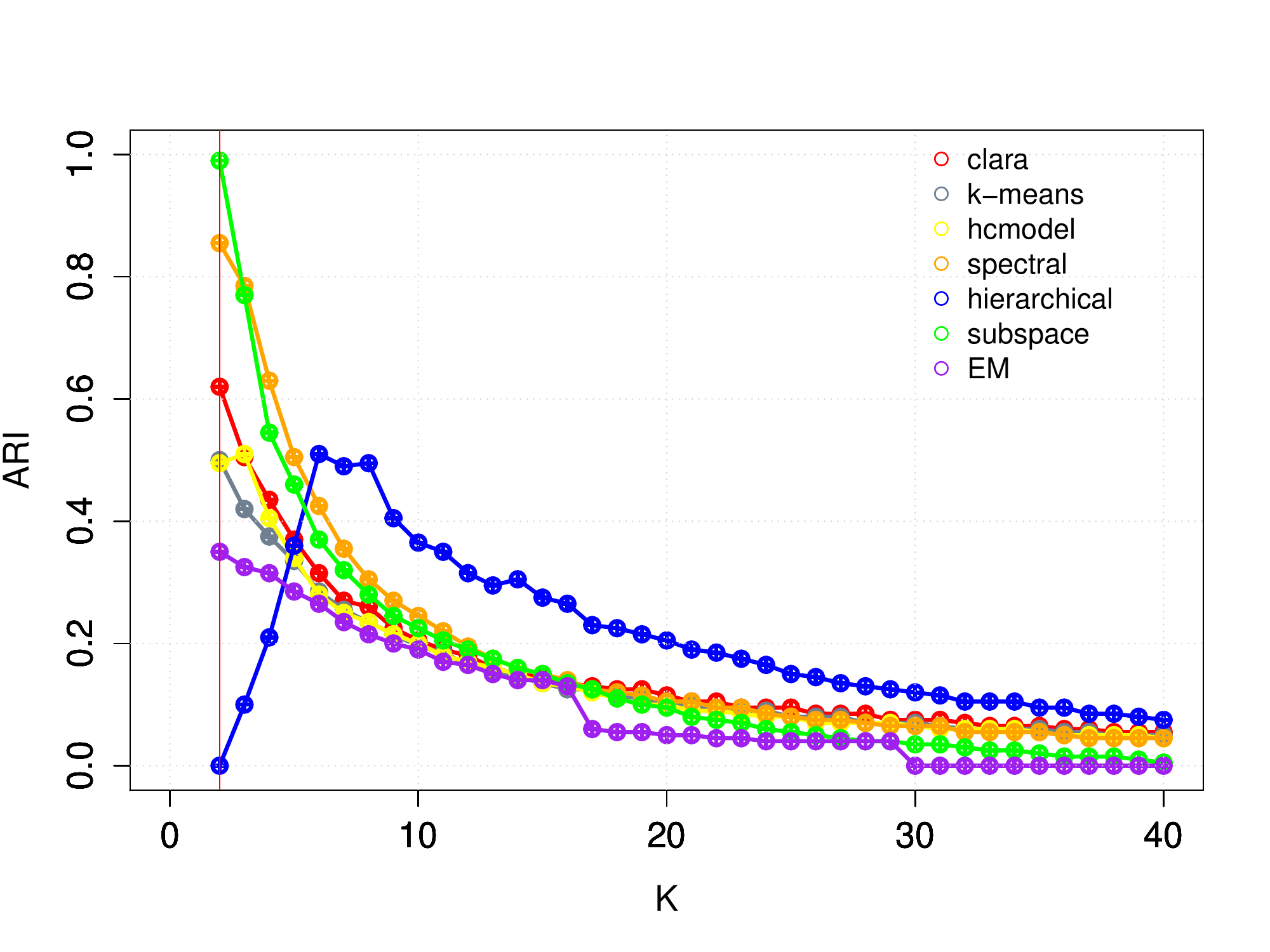}}\label{fig:clusteringKc}}
          \end{subfigure}
           \begin{subfigure}[][]{
	      {\includegraphics[width=0.45\columnwidth]{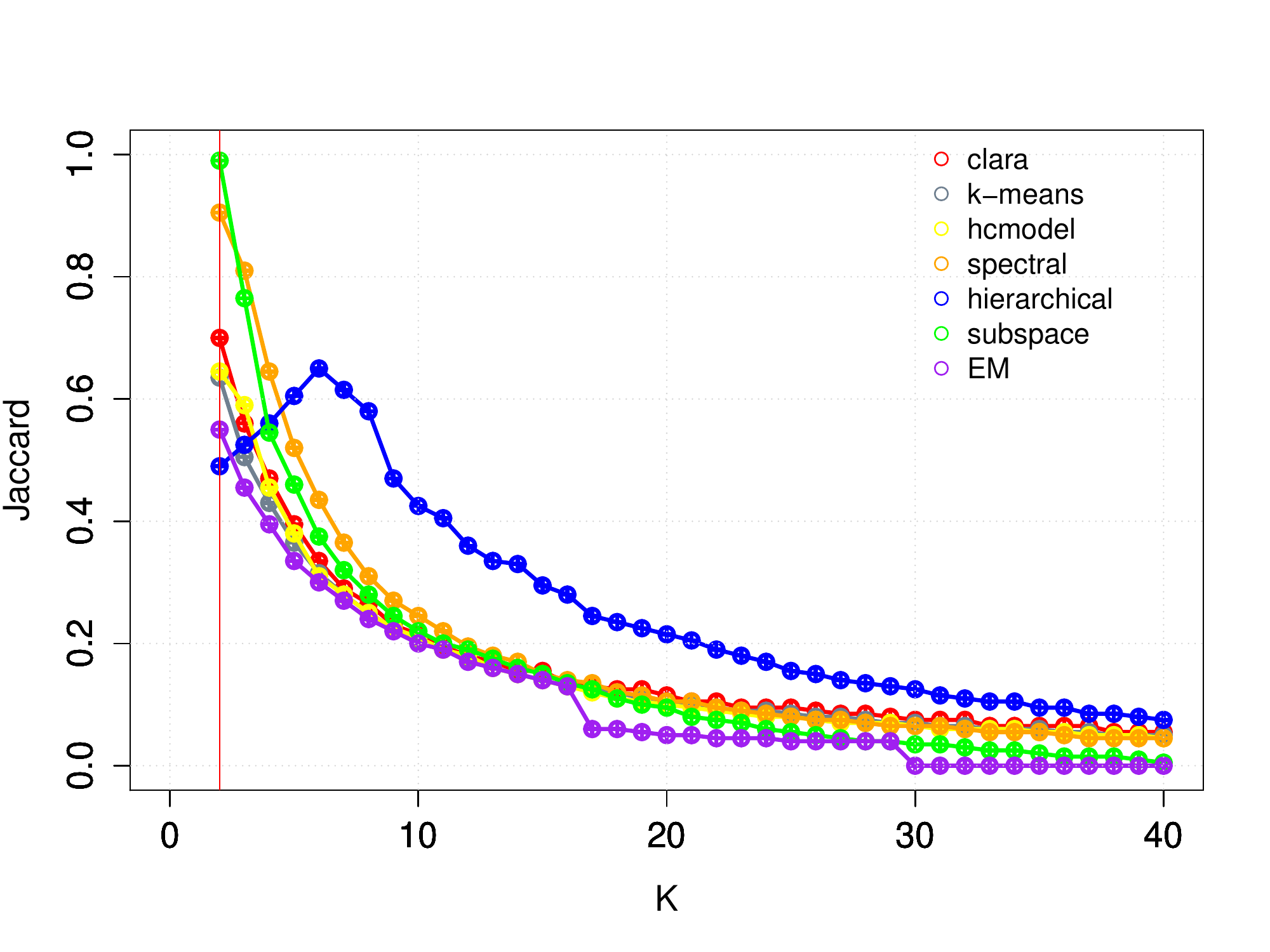}}\label{fig:clusteringKd}}
          \end{subfigure}  
   \end{center}
\caption{{\bf Performance of the algorithms when changing the expected number of clusters $K$ in the dataset.} Plots (a) and (b) correspond to the ARI and Jaccard indexes averaged for all datasets containing 10 classes and 10 features (DB10C10F). Plots (c) and (d) correspond to the same indexes  averaged for datasets containing 2 classes and 10 features (DB2C10F). The dashed red line indicates the actual number of clusters in the dataset.}
\label{fig:clusteringK}
\end{figure*}

It is also interesting to verify the performance of the clustering algorithms when setting distinct values for the expected number of classes $K$ in the dataset. Such a value is usually not known beforehand in real datasets. For instance, one might expect the data to contain 10 classes, and, as a consequence, set $K=10$ in the algorithm, but the objects may actually be better divided into 12 classes. An accurate algorithm should still provide reasonable results when setting a wrong number of classes. Considering this, we varied $K$ for each algorithm and verified the resulting variation in accuracy. In order to simplify the analysis, we consider a two-fold variation of datasets: (i)  a dataset comprising objects described by 10 features and divided into 10 classes (DB10C10F); and (ii) a dataset comprising objects described by 10 features and divided into 2 classes (DB2C10F). In Figures~\ref{fig:clusteringKa} and~\ref{fig:clusteringKb}, we show the average ARI and Jaccard indexes calculated for DB10C10F, while the same indexes for DB2C10F are shown in Figures~\ref{fig:clusteringKc} and~\ref{fig:clusteringKd}. The results for DB10C10F indicate that setting $K<10$ leads to a markedly worse performance than cases where $K>10$, which suggests that a slight overestimation of the number of classes does not lead to much worse performance. Therefore, a good strategy for choosing $K$ seems to be setting it to values that are slightly larger than the number of expected classes. 
An interesting behavior is observed for hierarchical clustering. The accuracy improves as the number of expected classes increases. This behavior is due to the default value of the \textit{method} parameter, which is set as ``average". The ``average'' value means that the unweighted pair group method with arithmetic mean  (UPGMA) is used to agglomerate the points. UPGMA is the average of the dissimilarities between the points in one cluster and the points in the other cluster. The poor performance of UPGMA in recovering the original groups, even with high subgroup differentiation, is because UPGMA tends to result in highly unbalanced clusters, that is, the majority of the objects are assigned to a few clusters while many other clusters contain only one or two objects.

The results obtained for the default parameters are summarized in Table~\ref{tab:resumo1}. The table is divided into four parts, each part corresponds to a performance metric. For each performance metric, the value in row $i$ and column $j$ of the table represents the average performance of the method in row $i$ minus the average performance of the method in column $j$. The last column of the table indicates the average performance of each algorithm. We note that the averages were taken over all generated datasets.

\begin{table}[h]
\caption{\label{tab:resumo1}{\bf Average difference of accuracies obtained when clustering algorithms are used with their default configuration of parameters.} In general, the spectral algorithm provides the highest accuracy rate among all evaluated methods.} 
%\centering
\scriptsize
%\resizebox{0.5\textwidth}{!}{\begin{minipage}{0.7\textwidth}
\begin{tabular}{|l|cccccccc|c| }

	\hline
	\multicolumn{1}{|c|}{\multirow{1}{*}{ }} & \multirow{1}{*}{\bf Algorithm} & \multirow{1}{*}{\bf hierarchical} & \multirow{1}{*}{\bf k-means}  & \multirow{1}{*}{\bf clara}  & \multirow{1}{*}{\bf spectral}   & \multirow{1}{*}{\bf hcmodel} &  \multirow{1}{*}{\bf subspace} & \multirow{1}{*}{\bf EM} & \multirow{1}{*}{\bf MAcc}    \\

	\hline
		   
	NMI & hierarchical	& -			& -22.68\%	 & -20.70\%   & -35.17\%  & -20.07\%  & -16.60\%  & -7.51 \%  & 45.53\%   \\
	& k-means		& 22.68\%		& -  & 1.98\%   & -12.49\%  &2.61\% & 6.08\%  & 15.17 \%  & 68.21\%    \\
	& clara		& 20.70\%	        &  -1.98\% &-  & -14.47\% &0.63\% & 4.10\%  & 13.19\% & 66.23\%    \\
	& spectral		& 35.17\%		& 12.49\%  & 14.47\%  & -  &15.10\% & 18.57\%  & 27.66 \%  & 80.70\%  \\
	& hcmodel		& 20.07\%		& -2.61\%  &-0.63 \% &-15.10\%  & - & 3.47\%  & 12.56\%   & 65.60\%  \\ 	
	& subspace		& 16.60\%		& -6.08\%  & -4.10\% &-18.57\%  & -3.47   & - & 9.09\%  & 62.13\%  \\ 	
	& EM			& 7.51\%		& -15.17\%  & -13.19\% &-27.66\%  & -12.56   & -9.09\% & -  & 53.04\% \\ 	
	\hline
		
	ARI & hierarchical	& -			& -27.38\%	 & -25.03\%   & -40.99\% &-18.72\% & -33.77\% & -9.27 \% &  24.81\%    \\
	& k-means		& 27.38\%		&-               &2.35\%      & -13.61\%     & 8.66\%  & -6.39\% & 18.11\%  &  52.19\%   \\
	& clara		& 25.03\%	        & -2.35\%        & -          & -15.96\% & 6.31\% & -8.74\% & 15.76\%  & 49.84\%    \\
	& spectral		& 40.99\%		& 13.61\%        & 15.96\%    &  -  & 22.27\% & 7.22\% & 31.72\%   & 65.80\%  \\ 
	& hcmodel		& 18.72\%		& -8.66\%  	& -6.31\%   & -22.27\% & -  & -15.05\% & 9.45\%  & 43.53\%  \\ 
	& subspace		& 33.77\%		& 6.39\%  & 8.74\% & -7.22\%  & 15.05   & - & 24.50\% & 58.58\%   \\ 	
	& EM			& 9.27\%		& -18.11\%  & -15.76\% & -31.72\%  & -9.45   & -24.50\% & - & 34.08\%   \\

	\hline
	Jaccard & hierarchical	& -			& -14.45\%	 &  -11.78\%   &  -26.98\% & -8.50\% & -23.53\% & -2.58\%  & 33.24\%     \\
	& k-means		& 14.45\%		&-               & 2.67\% & -12.53\%      & 5.95\% & -9.08\% & 11.87\%    & 47.69\%     \\
	& clara		& 11.78\%		& -2.67\%        &-            & -15.20\% &3.28\% & -11.75\% & 9.20\%   & 45.02\%  \\
	& spectral		& 26.98\%		& 12.53\%        & 15.20\%      & - &18.48\%    & 3.45\% & 24.40\%       & 60.22\% \\ 
	& hcmodel		& 8.50\%		& -5.95\%  	& -3.28\%  & -18.48 & - & -15.03\% & 5.92\%   & 41.74\%  \\ 	
	& subspace		& 23.53\%		& 9.08 \%  &11.75 \% & -3.45\%  & 15.03   & - & 20.95\% & 56.77\%  \\ 	
	& EM			& 2.58\%		& -11.87\%  & -9.20\% & -24.40\%  & -5.92   & -20.95\% & - & 35.82\%  \\

	\hline
	FM & hierarchical	& -			& -11.66\%	 &  -8.29\%    &  -21.87\% &-6.61\% & -18.31\% &-0.19 \%   & 51.40\%    \\
	& k-means		& 11.66\%		& -              & 3.37\%       & -10.21\%  &5.05\% & -6.65\% & 11.47\%  & 63.06\%   \\
	& clara		& 8.29\%		& -3.37\%        & -            & -13.58\% &1.68\% & -10.02\% & 8.10\%  & 59.69\%   \\
	& spectral		& 21.87\%		& 10.21\%        & 13.58\%     &  -  &15.26\%   & 3.56\% & 21.68\%     & 73.27\% \\ 
	& hcmodel		& 6.61\%		& -5.05\%  	& -1.68\%  & -15.26 \% & - & -11.70\% & 6.42\%   & 58.01\%  \\ 
	& subspace		& 18.31\%		& 6.65\%  & 10.02\% & -3.56\%  & 11.70   & - & 18.12\% & 69.71\%  \\ 	
	& EM			& 0.19\%		& -11.47\%  & -8.10\% & -21.68\%  & -6.42   & -18.12\% & - & 51.59\%  \\ 	
	
	\hline  

\end{tabular}

%\end{minipage}}

\end{table}

The results shown in Table~\ref{tab:resumo1} indicate that the spectral algorithm tends to outperform the other algorithms by at least 10\%. On the other hand, the hierarchical method attained a poor performance in all considered cases. Another interesting result is that the k-means, clara and hcmodel provide equivalent performance when considering all datasets. In light of the results, we can conclude that the spectral method is to be preferred when no optimitization of parameters values is performed.

\subsection{One-dimensional analysis} 
\label{s:resonedim}

The objective of the one-dimensional
analysis is to verify how sensitive the accuracy of the clustering algorithms is to the variation of a single parameter.
In addition, this analysis is also useful to verify if a very simple optimization strategy can lead to significant improvements in performance.
For the one-dimensional analysis, we considered the databases DB2C2F and DB10C2F with $\alpha=2.5$ and $\alpha=4.3$, respectively. We also considered DB2C10F and DB10C10F with $\alpha=1.16$ and $\alpha=1.75$, respectively. 
For each parameter, we varied its values while keeping the other parameters value in their default configuration. The effect of varying the values of a single parameter $P$ was quantified by comparing the obtained accuracy $\Gamma(x)$ when the parameter takes the value $x$ and the accuracy $\Gamma_{\textrm{def}}$ achieved with the default configuration of parameters. The improvement in performance was quantified in terms of the average ($\langle S \rangle$) and maximum value ($\max S$), given by
\begin{equation}
	\langle S \rangle = \frac{1}{n_P} \sum_x \Big{(} \Gamma(x) - \Gamma_{\textrm{def}}\Big{)},
\end{equation}
\begin{equation}
	\max S = \max_x \left(\Gamma(x) - \Gamma_{\textrm{def}}\right),
\end{equation}
where $n_P$ is the cardinality of all possible values taken by the parameter $P$ in our experiments. We also measured the sensitivity of varying the values of $P$ using the standard deviation $\Delta S$:
\begin{equation}
	\Delta S = \Bigg{[} \frac{1}{n_P} \sum_x \Big{(}\Gamma(x)-\Gamma_{\textrm{def}}-\langle S \rangle\Big{)} ^2\Bigg{]}^{1/2}.
\end{equation}
In addition to the aforementioned quantities, we also measured, for each dataset, the maximum accuracy obtained when varying each single parameter of the algorithm. We then calculate the average of maximum accuracies, $\langle\max \textrm{Acc}\rangle$, obtained over all considered datasets.
In Table \ref{tab:oneD2}, we show the values of $\langle S \rangle$, $\max S$, $\Delta S$ and $\langle\max \textrm{Acc}\rangle$ for datasets containing two features. When considering a two-class problem (DB2C2F), a significant improvement in performance ($\langle S\rangle=10.75\%$) was observed when varying parameter \textit{modelName} of the EM method. Similarly, parameter \textit{kpar} of the spectral method provided an average improvement of $\langle S\rangle = 7.36\%$. For all other cases, only minor average gain in performance was observed. For the 10-class problem, we notice that an inadequate value for parameter \textit{method} of the hierarchical algorithm can lead to much worse accuracy ($-16.15\%$ on average). In most cases, however, the average variation in performance was small.

\begin{table*}[htbp]
\caption{\label{tab:oneD2}{\bf One-parameter analysis performed in DB2C2F and DB10C2F.} This analysis is based on the performance (measured through the ARI index) obtained when varying a single parameter of the clustering algorithm, while maintaining the others in their default configuration. $\langle S \rangle$, $\max S$, $\Delta S$ are associated with the average, standard deviation and maximum difference between the performance obtained when varying a single parameter and the performance obtained for the default parameter values. We also measure $\langle\max \textrm{Acc}\rangle$, the average of best ARI values obtained when varying each parameter, where the average is calculated over all considered datasets.}
\centering
%\scriptsize
 \resizebox{1.00\textwidth}{!}{\begin{minipage}{\textwidth}
\begin{tabular}{@{} |lc|cccc|cccc|}
	\cline{3-10}
	\multicolumn{2}{c|}{} & \multicolumn{4}{c}{DB2C2F} & \multicolumn{4}{c|}{DB10C2F} \\
	\hline
	\multirow{2}{*}{\bf Algorithm} & \multirow{2}{*}{\bf Parameter} & $\langle S \rangle$ & $\Delta S$ & $\max S$  & $\langle\max \textrm{Acc}\rangle$ & $\langle S \rangle$ & $\Delta S$ & $\max S$ & $\langle\max \textrm{Acc}\rangle$  \\
 	& & (\%) & (\%) & (\%) & (\%) & (\%) & (\%) & (\%) & (\%)    \\
	\hline
	k-means         & iter.max &  0.05 & 2.37  & 14.46 & 51.5 & 0.04 & 0.91 & 4.49 & 47.3  \\
	k-means         & nstart &   1.98 & 5.62  & 16.73 & 51.9 & 1.24& 1.98 & 6.80 & 47.9  \\
	k-means         & algorithm &  0.29  & 2.46  & 6.63 & 49.8 & -0.92 & 1.29 & 0.65 & 45.0  \\ 
	clara & metric  & -1.52 & 8.10   & 11.27  & 49.6 & -3.66 & 5.36 & 5.10 & 42.5   \\
	clara & samples  & -0.10 & 3.82  & 6.39 & 52.3 & -0.21 & 3.03 & 7.48 &  47.5  \\
	clara & sampsize  & -2.78 & 12.96  & 27.31 & 54.0  & -0.54 & 2.92 & 4.88 & 47.1 \\
	clara & rngR  & -0.16 & 3.19  & 4.19 & 51.0 & -4.53 & 4.04 & -0.03 & 41.7\\
	hierarchical & metric  & 5.27 & 22.28  & 63.65 & 23.3  & 1.83 & 3.64 & 9.26 & 42.3   \\
	hierarchical & method  & 2.07 & 36.90  & 100.0 & 57.2   & -16.15 & 21.26 & 15.89 & 46.5 \\
	hierarchical & par.method & 0.0 & 0.0  & 0.0 & 18.0& 0.0& 0.0 & 0.0 & 40.5 \\
	spectral & kernel  & -0.61 & 10.42 & 39.45 & 43.7  & -0.3 & 2.84 & 6.78 & 48.0 \\
	spectral & kpar & 7.36 & 16.78  & 33.3 & 44.6  & -1.83 & 3.16 & 3.35 & 43.5   \\
	spectral & iter & 1.14 & 19.19  & 85.34 & 54.1 & 0.06 & 2.62 & 5.84 & 47.9  \\
	spectral & mod.simple & -2.11 & 9.7  & 33.32 & 43.0 &0.54 & 2.02 & 4.5 & 47.7   \\
	hcmodel & modelName &  -2.41 & 19.48 & 29.89 & 60.0  & -0.56 & 3.26 & 6.44 & 48.4 \\
	hcmodel & use & -2.14 & 10.14  & 12.58 & 57.4 & -0.50 & 1.11 &2.19 & 47.5  \\
	EM  &z &  1.71& 8.77  & 19.34 & 33.9 & 7.04 & 8.30  &28.17 &  45.4 \\
	EM  &modelName & 10.75 &  26.18 &66.64 & 64.4 & 0.14 & 6.25  & 16.20 & 45.0\\
	\hline
\end{tabular}
\end{minipage}}
\end{table*}

In Table \ref{tab:oneD1}, we show the values of $\langle S \rangle$, $\max S$, $\Delta S$ and $\langle \max \textrm{Acc}\rangle$ for datasets described by 10 features. For the the two-class clustering problem, a moderate improvement can be observed for the k-means algorithm through the variation of parameter \emph{nstart}. A significant increase in accuracy was observed when varying parameter \emph{method} of the hierarchical algorithm and parameter \emph{modelName} of the EM method. These parameters provided an average accuracy gain of $8.76\%$ and $18.8\%$, respectively. A similar behavior was obtained when the number of classes was set to $C=10$. For 10 classes, the variation of \emph{method} in the hierarchical algorithm provided an average improvement of $6.72\%$.  A high improvement was also observed when varying parameter \emph{modelName} of the EM algorithm, with an average improvement of $13.63\%$.  

Differently from the parameters discussed so far, the variation of some parameters plays a minor role in the discriminative power of the clustering algorithms. This is the case, for instance, of parameters \emph{kernel} and \emph{iter} of the spectral clustering algorithm and parameter \emph{iter.max} of the kmeans clustering. In some cases, the effect of a unidimensional variation of parameter resulted in reduction of performance. For instance, the variation of \emph{min.individuals} and \emph{models} of the subspace algorithm provided an average loss of accuracy on the order of $\langle S \rangle = -20\%$, depending on the dataset. A similar trend was observed for parameters \textit{metric} and \textit{rngR} of the clara algorithm.

\begin{table*}[htbp]
\caption{\label{tab:oneD1}{\bf One-parameter analysis performed in DB2C10F and DB10C10F.} This analysis is based on the performance obtained when varying a single parameter, while maintaining the others in their default configuration. $\langle S \rangle$, $\max S$, $\Delta S$ are associated with the average, standard deviation and maximum difference between the performance obtained when varying a single parameter and the performance obtained for the default parameter values. We also measure $\langle\max \textrm{Acc}\rangle$, the average of best ARI values obtained when varying each parameter, where the average is calculated over all considered datasets.}
%\centering
%\scriptsize
 \resizebox{1.0\textwidth}{!}{\begin{minipage}{\textwidth}
\begin{tabular}{@{} |lc|cccc|cccc|}
	\cline{3-10}
	\multicolumn{2}{c|}{} & \multicolumn{4}{c}{DB2C10F} & \multicolumn{4}{c|}{DB10C10F} \\
	\hline
	\multirow{2}{*}{\bf Algorithm} & \multirow{2}{*}{\bf Parameter} & $\langle S \rangle$ & $\Delta S$ & $\max S$  & $\langle\max \textrm{Acc}\rangle$ &  $\langle S \rangle$ & $\Delta S$ & $\max S$ & $\langle\max \textrm{Acc}\rangle$  \\
 	& & (\%) & (\%) & (\%) & (\%) & (\%) & (\%) & (\%) & (\%)   \\
	\hline
	k-means         & iter.max & 0.30  & 8.13  &36.36 & 53.2 & 0.14 & 1.92 & 6.41 & 56.6  \\
	k-means         & nstart & 5.26   & 12.0 & 36.36 & 53.5 & 2.68 & 2.65 & 9.43 & 57.8 \\
	k-means         & algorithm & -0.35   & 6.72  & 25.5 & 42.3  & -2.11 & 3.3 & 2.71 & 52.7  \\ 
	clara & metric  & -10.9  &22.31  & 25.05 & 51.8  &  -16.63 & 6.84  & -5.1  & 37.6   \\
	clara & samples  & 1.04 & 8.94  & 25.05 & 60.4 & -4.83   & 8.96  & 10.26  & 51.9   \\
	clara & sampsize  & 0.44 & 13.94  & 37.31 & 61.0 & -4.46  & 9.97 & 14.18  & 57.6     \\
	clara & rngR  & -2.89 & 15.08  & 25.05 & 56.9 & -14.75  & 6.29  & -5.21 & 39.3  \\
	hierarchical & metric  & 4.82 & 21.46 & 96.0 & 9.7  & 1.15 & 8.52 & 27.18 & 19.2     \\
	hierarchical & method  & 8.76 & 21.93 & 100.0 & 43.7 & 6.72 & 25.52 & 71.1 & 61.5    \\
	hierarchical & par.method & 0.00 & 0.00 & 0.00 & 0.00  & 0.0 & 0.0 & 0.0 & 13.8    \\
	spectral & kernel  &0.64  &15.91  &50.56 & 87.9  & 1.3 & 7.13 & 15.81 & 82.3  \\
	spectral & kpar &-1.08  &16.88  & 50.56 & 88.1  & -2.25 & 5.81 & 6.03 & 71.7     \\
	spectral & iter & -0.96 &15.91  &50.56 & 87.9 & 0.45 & 7.27 & 20.01 & 79.8  \\
	spectral & mod.simple & 3.36 &15.72  &50.56 & 87.9  & -1.35 & 7.55 & 14.24 & 78.7  \\
	subspace & models & -1.77 & 36.80  & 97.4 & 100.0 & -22.44 & 8.92 & -6.7 & 69.6  \\
	subspace & init & -0.78 & 23.47   & 97.4 & 99.5 & -0.57  & 9.29  & 11.13 & 87.4    \\
	subspace & algo & -1.32 & 1.99  & 0.27 & 88.9 & 0.7 & 1.1 &  1.9 &  87.4  \\
	subspace & min.individuals & -26.9 & 43.17  & 10.73 & 90.9& -12.32& 16.6  & 7.78 & 89.1 \\ 
	hcmodel & modelName & 3.70 & 24.23 & 75.6 & 51.3 & 3.63 &4.89 & 14.4 & 61.5  \\
	hcmodel &use & -0.92 &17.68 &51.47 & 49.1  & -1.86 & 6.09 & 10.69 & 55.9  \\
        EM &z & 1.68& 8.62 &18.99 & 29.9 & -0.35 & 5.49& 15.06 & 43.3   \\
        EM &modelName & 18.80& 31.93 & 96.62 & 100.0 & 13.63 & 16.09 & 64.52 & 91.6 \\
    
	\hline
\end{tabular}
\end{minipage}}
\\ \ \\
\end{table*}

\subsection{Multi-dimensional analysis}

A complete analysis of the performance of a clustering algorithm requires the simultaneous variation of all of its parameters. Nevertheless, such a task is difficult to do in practice, given the large number of parameter combinations that need to be taken into account. Therefore, here we consider a random variation of parameters aimed at obtaining a sampling of each algorithm performance for its complete multi-dimensional parameter space.

The methodology is applied as follows. Considering the one-dimensional variation of parameters, presented in the previous section, we identify the parameter bounds, $[P_{\text{min}},P_{\text{max}}]$, where the classification either does not significantly change anymore or provides much worse results when compared to the default parameter value. Such bounds define the interval where the parameter will be randomly sampled. In order to generate the values for a given set of parameters $P^{(1)},P^{(2)},\dots,P^{(n)}$ of an algorithm, we randomly sample each parameter $P^{(i)}$ according to a uniform distribution defined in the interval $[P_{\text{min}}^{(i)},P_{\text{max}}^{(i)}]$. This procedure generates sets of parameter values, which are then used to evaluate the performance of the algorithms. For each algorithm, 500 sets of parameters were generated.

\begin{figure*}[ht!]
    \begin{center}
	      \includegraphics[width=0.8\linewidth]{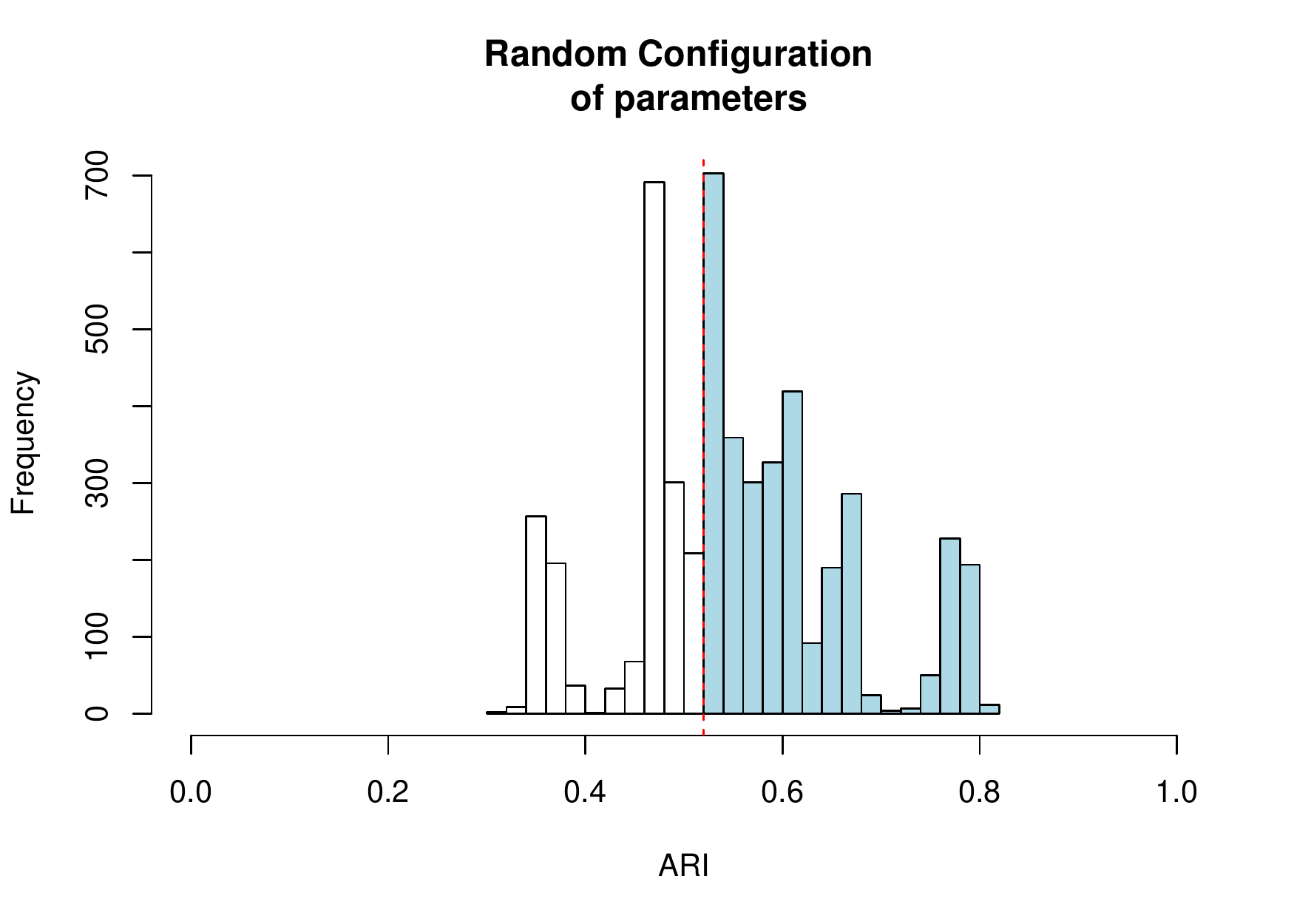}\label{fig:rand10CA}
   \end{center}
\caption{{\bf Distribution of ARI values obtained for the random sampling of the k-means parameters.} The algorithm was applied to dataset DB10C10F, and 500 sets of parameters were drawn.}
\label{fig:acuracyRandomParam1}
\end{figure*}

The performance of the algorithms for the different sets of parameters was evaluated according to the following procedure. Consider the histogram of ARI values obtained for the random sampling of parameters for the k-means algorithm, shown in Figure~\ref{fig:acuracyRandomParam1}. The red dashed line indicates the ARI value obtained for the default parameters of the algorithm. The light blue shaded region indicates the parameters configurations where the performance of the algorithm improved. From this result we calculated four main measures. The first, which we call p-value, is given by the area of the blue region divided by the total histogram area, multiplied by 100 in order to result in a percentage value. The p-value represents the percentage of parameter configurations where the algorithm performance improved when compared to the default parameters configuration. The second, third and fourth measures are given by the mean, $\langle R\rangle$, standard deviation, $\Delta R$, and maximum value, $\max R$, of the relative performance for all cases where the performance is improved (e.g. the blue shaded region in Figure~\ref{fig:acuracyRandomParam1}). The relative performance is calculated as the difference in performance between a given realization of parameter values and the default parameters. The mean indicates the expected improvement of the algorithm for the random variation of parameters. The standard deviation represents the stability of such improvement, that is, how certain one is that the performance will be improved when doing such random variation. The maximum value indicates the largest improvement obtained when random parameters are considered. We also measured the average of the maximum accuracies $\langle\max\textrm{ARI}\rangle$ obtained for each dataset when randomly selecting the parameters. In Section S2 of the supplementary material we show the distribution of ARI values obtained for the random sampling of parameters for all clustering algorithms considered in our analysis.  

In Table \ref{tab:multD3} we show the performance (ARI) of the algorithms for dataset DB2C2F when applying the aforementioned random selection of parameters. The EM method is the only algorithm with a p-value larger than 50\%. Also, a high average gain in performance was observed for the EM (22.1\%) and hierarchical (30.6\%) algorithms.
Moderate improvement was obtained for the hcmodel, kmeans and spectral algorithms.

\begin{table*}[htbp]
  \caption{\label{tab:multD3}{\bf Multi-parameter analysis performed in dataset DB2C2F.} The p-value represents the probability that the classifier set with a random configuration of parameters outperform the same classifier set with its default parameters. $\langle R\rangle$, $\Delta R$ and $\max R$ represent the average, standard deviation and maximum value of the improvement obtained when random parameters are considered. Column $\langle\max\textrm{ARI}\rangle$ indicates the average of the best accuracies obtained for each dataset.}
  \centering
 % \scriptsize
  \begin{tabular}{@{} lccccc}
    \toprule
     \multirow{2}{*}{\bf Algorithm} & \multirow{1}{*}{p-value} & \multirow{1}{*}{$\langle R\rangle$}  & \multirow{1}{*}{$\Delta R$} & \multirow{1}{*}{$\max R$} & \multirow{1}{*}{$\langle\max\textrm{ARI}\rangle$} \\ 
     & (\%)& (\%) & (\%) & (\%) &  (\%) \\
    \hline
    EM & 68.1 &  22.1 & 21.7  & 69.6  & 69.0   \\
    hierarchical	& 43.9   & 30.6  &  33.6  & 100.0  & 63.0   \\
    clara	& 29.2   & 4.9   & 4.7 & 27.3 &  60.0 \\
    hcmodel    	& 25.8    & 13.3 & 8.2   & 29.9  & 63.0 \\ 
    k-means	&  21.7   &  13.2 &  3.9   & 21.4 &   55.0   \\
    spectral	&  47.0 	 & 14.7 & 13.9 & 85.3    &   59.0 \\ 
    \hline
     \hline

  \end{tabular}
  % \end{minipage}}
\end{table*}

The performance of the algorithms for dataset DB10C2F is presented in Table \ref{tab:multD5}. A high p-value was obtained for the EM (76.5\%) and k-means (77.7\%) algorithms. Nevertheless, the average improvement in performance was relatively low for all algorithms. 

\begin{table*}[htbp]
  \caption{\label{tab:multD5}{\bf Multi-parameter analysis performed in dataset DB10C2F.} The p-value represents the probability that the classifier set with a random configuration of parameters outperform the same classifier set with its default parameters. $\langle R\rangle$, $\Delta R$ and $\max R$ represent the average, standard deviation and maximum value of the improvement obtained when random parameters are considered. Column $\langle\max\textrm{ARI}\rangle$ indicates the average of the best accuracies obtained for each dataset.}
  \centering
 \begin{tabular}{@{} lccccc}
    \toprule
     \multirow{2}{*}{\bf Algorithm} & \multirow{1}{*}{p-value} &   \multirow{1}{*}{$\langle R\rangle$}  & \multirow{1}{*}{$\Delta R$} & \multirow{1}{*}{$\max R$} & \multirow{1}{*}{$\langle\max \textrm{ARI}\rangle$} \\ 
& (\%)& (\%) & (\%) & (\%) & (\%) \\
    \hline
   EM & 76.5 & 7.5 & 8.0  & 35.8 & 51.4   \\
   clara & 54.7 & 2.3 & 1.8   & 9.0 & 51.0  \\
   k-means & 77.7  & 2.2  & 1.7 & 6.9  & 49.0 \\
   hcmodel & 28.4 & 2.7 & 2.5   & 6.8 & 49.0  \\  
   hierarchical	& 36.6  & 5.9  & 4.2  & 21.7  & 49.0  \\
   spectral	& 40.0   & 2.3  & 1.6  & 8.0  & 52.0  \\  
   \hline    \hline
      \end{tabular}
   %   \end{minipage}}
\end{table*}

A more diverse variation in performance was observed for dataset DB2C10F, with results shown in Table~\ref{tab:multD2}. The EM, kmeans and hierarchical clustering are the only algorithms with a p-value larger than 50\%. In such cases, when the performance was improved, the average gain in performance was 30.1\%, 18.0\% and 25.9\%, respectively.
This means that the random variation of parameters might represent a valid approach for improving these algorithms. Actually, with the exception of clara, all methods display significant average improvement in performance for this dataset.
The results also show that a maximum accuracy of 100\% can be achieved for the EM and subspace algorithms.
\begin{table*}[htbp]
  \caption{\label{tab:multD2}{\bf Multi-parameter analysis performed in dataset DB2C10F.} The p-value represents the probability that the classifier set with a random configuration of parameters outperform the same classifier set with its default parameters. $\langle R\rangle$, $\Delta R$ and $\max R$ represent the average, standard deviation and maximum value of the improvement obtained when random parameters are considered. Column $\langle\max\textrm{ARI}\rangle$ indicates the average of the best accuracies obtained for each dataset.}
  \centering
 % \scriptsize
  \begin{tabular}{@{} lccccc}
    \toprule
     \multirow{2}{*}{\bf Algorithm} & \multirow{1}{*}{p-value} & \multirow{1}{*}{$\langle R\rangle$}  & \multirow{1}{*}{$\Delta R$} & \multirow{1}{*}{$\max R$} & \multirow{1}{*}{$\langle\max\textrm{ARI}\rangle$} \\ 
     & (\%)& (\%) & (\%) & (\%) &  (\%) \\
    \hline
    EM & 70.8 & 30.1  & 29.9  & 96.6  & 100.0  \\
    hierarchical	& 52.0   &25.9  & 31.4   & 100.0  & 80.0   \\
    subspace  &    11.1   &  43.1   &  45.4 & 97.4	& 100.0    \\ 
    clara	& 44.9   & 6.5   & 6.3 & 37.3 & 70.0  \\
    hcmodel    	& 38.4    & 31.8 & 25.3  & 81.2  & 70.0 \\ 
    k-means	& 50.1    & 18.0  & 7.1    & 62.4 & 60.0     \\
    spectral	&  48.9 	 & 9.9 & 18.5 &  31.5   & 90.0   \\ 
    \hline
    \hline

  \end{tabular}
  % \end{minipage}}
\end{table*}

In Table \ref{tab:multD4} we show the performance of the algorithms for dataset DB10C10F. The p-value for the EM, clara and k-means indicates that the random selection of parameters usually improves the performance of these algorithms, although only the EM method display a significant improvement in performance ($\langle R\rangle = 17.1\%$). The hierarchical algorithm can be significantly improved by the considered random selection of parameters. This is a consequence of the default value of parameter \emph{method}, which, as discussed in Section~\ref{s:resonedim}, is not appropriate for this dataset. 

\begin{table*}[htbp]
  \caption{\label{tab:multD4}{\bf Multi-parameter analysis performed in dataset DB10C10F.} The p-value represents the probability that the classifier set with a random configuration of parameters outperform the same classifier set with its default parameters. $\langle R\rangle$, $\Delta R$ and $\max R$ represent the average, standard deviation and maximum value of the improvement obtained when random parameters are considered. Column $\langle\max\textrm{ARI}\rangle$ indicates the average of the best accuracies obtained for each dataset.}
  \centering
 \begin{tabular}{@{} lccccc}
    \toprule
     \multirow{2}{*}{\bf Algorithm} & \multirow{1}{*}{p-value} &   \multirow{1}{*}{$\langle R\rangle$}  & \multirow{1}{*}{$\Delta R$} & \multirow{1}{*}{$\max R$} & \multirow{1}{*}{$\langle\max \textrm{ARI}\rangle$} \\ 
& (\%)& (\%) & (\%) & (\%) & (\%) \\
    \hline
   EM &  86.0& 17.1 & 15.5  & 69.1 & 100.0   \\
   clara & 72.1 & 7.1 & 4.4  & 22.8 & 68.0 \\
   k-means & 83.0  & 4.3 & 2.3 & 12.0 & 60.0\\
   hcmodel & 53.4  &7.4  & 4.6  & 17.5 & 64.0  \\  
   hierarchical	& 51.9  &32.1   &19.4   &72.9 & 68.0  \\
   spectral	&49.1   &5.6  &4.1  & 19.7 & 87.3  \\ 
   subspace &  10.7 & 7.5   &4.7   & 21.4 & 99.3 \\  
   \hline
   \hline
      \end{tabular}
   %   \end{minipage}}
\end{table*}

\section{Conclusions}

Clustering data is a complex task involving the choice between many different methods, parameters and performance metrics, with implications in many real-world problems~\cite{deArruda20126174,10.1371/journal.pone.0157988,1742-5468-2015-3-P03005,10.1371/journal.pone.0162259,10.1371/journal.pone.0156576,Colavizza20161037,0295-5075-19-3-015}.  Consequently, the analysis of the advantages and pitfalls of clustering algorithms is also a difficult task that has been received much attention. Here, we approached this task focusing on a comprehensive methodology for generating a large diversity of heterogeneous datasets with precisely defined properties such as the distances between classes and correlations between features. Using packages in the R language, we developed a comparison of the performance of seven popular clustering methods applied to 270 artificial datasets. Three situations were considered: default parameters, single parameter variation and random variation of parameters. Besides serving as a practical guidance to the application of clustering methods when the researcher is not an expert in data mining techniques, a number of interesting results regarding the considered clustering methods were obtained. 

Regarding the default parameters, the difference in performance of clustering methods was not significant for low-dimensional datasets. Specifically, the Kruskal-Wallis test on the differences in performance when 2 features were considered resulted in a p-value of $p = 0.07$ (with a chi-squared distance of $\chi^2=10.26$). When more features were considered, the performance became markedly distinct among the methods. Considering 50 features resulted in a p-value of $p = 1.4\times 10^{-6}$ for the Kruskal-Wallis test ($\chi^2 = 37.48$) in the difference in performance.

The Spectral method provided the best performance when using default parameters, with an Adjusted Rand Index (ARI) of 65.80\%, as indicated in Table~\ref{tab:resumo1}. In contrast, the hierarchical method showed the worst results with an ARI of 24.81\%. On the other hand, the hierarchical clustering based on parametric Gaussian mixture models, implemented in the function hc from the \textit{mclust} package, had a better performance with an ARI of 43.53\%.  It is also interesting that underestimating the number of classes in the dataset led to worse performance than in overestimation situations. This was observed for all algorithms and is in accordance with previous results~\cite{erman2006traffic}.  

Regarding single parameter variations, for datasets containing 2 features, only the hierarchical and EM methods showed significant performance variation. On the other hand, for datasets containing 10 features, most methods could be readily improved through changes on selected parameters.

With respect to the multidimensional analysis for datasets containing two classes and ten features, the EM, hcmodel, subspace and hierarchical algorithm showed significant gain in performance. The EM algorithm also resulted in a high p-value (70.8\%), which indicates that many parameter values for this algorithm can provide better results than the default configuration. For datasets containing ten classes and ten features, the improvement was significantly lower for almost all the algorithms, with the exception of the hierarchical clustering.
For datasets containing ten classes and two features, the performance of the algorithms for the multidimensional selection of parameters was similar to the performance when using the default parameters. This suggests that the algorithms are not sensitive to parameter variations for this dataset. 

In Tables~\ref{tab:ranking2} and~\ref{tab:ranking1} we show a summary of the best accuracies obtained during our analysis. The tables contain the best performance, measured as the ARI of the resulting partitions, achieved by each algorithm in the three considered situations (default, one- and multi-dimensional adjustment of parameters). The results are respective to datasets DB2C2F, DB10C2F, DB2C10F and DB10C10F. We observe that, for datasets containing 2 features, all algorithms show similar performance. For datasets containing 10 features, the subspace algorithm seems to consistently provide the best performance, although the EM algorithm can also achieve similar performance with some tuning of its parameters.

\begin{table*}[htbp]
\caption{\label{tab:ranking2}{\bf Summary table for the performance of clustering algorithms in datasets DB2C2F and DB10C2F}. $ARI_{def}$ represents the average accuracy obtained when considering the default parameters of the algorithms. $ARI_{best_p}$ represents the average of the best accuracies obtained when varying a single parameter. $ARI_{best_r}$ represents the average of the best accuracies obtained when parameters are randomly selected.}
\centering
%\scriptsize
\resizebox{1.0\textwidth}{!}{\begin{minipage}{\textwidth}
\begin{tabular}{@{} |l|ccc|ccc|}
\cline{2-7}
 \multicolumn{1}{l|}{}  & \multicolumn{3}{c}{DB2C2F} & \multicolumn{3}{c|}{DB10C2F} \\
\hline
 \multirow{2}{*}{\bf Algorithm} & 
\multirow{1}{*}{\bf$ARI_{def}$} & \multirow{1}{*}{\bf$ARI_{best_p}$} & \multirow{1}{*}{\bf $ARI_{best_r}$} & 
\multirow{1}{*}{\bf$ARI_{def}$} & \multirow{1}{*}{\bf$ARI_{best_p}$} & \multirow{1}{*}{\bf $ARI_{best_r}$} \\ 
 & (\%) & (\%) & (\%) & (\%) & (\%) & (\%) \\

\hline
 
EM     	  & 32.2  & 64.4 & 69.0 & 38.4 & 45.4  & 51.4 \\
spectral	  & 37.2 &  54.1 &  59.0 & 45.4 & 48.0 & 52.0  \\
clara	  & 51.1  & 54.0  & 60.0  & 46.2 & 47.5	& 51.0 \\
hcmodel      & 54.0 & 60.0  & 63.0  & 47.1 &  48.4	&  49.0\\
k-means      & 48.1 & 51.9  & 55.0  & 45.2 & 47.9	& 49.0 \\
hierarchical & 18.0  & 57.2  & 63.0  & 40.5 & 46.5	& 49.0 \\	 
\hline
   
\end{tabular}
\end{minipage}}
\end{table*} 

\begin{table*}[htbp]
\caption{\label{tab:ranking1}{\bf Summary table for the performance of clustering algorithms in datasets DB2C10F and DB10C10F}. $ARI_{def}$ represents the average accuracy obtained when considering the default parameters of the algorithms. $ARI_{best_p}$ represents the average of the best accuracies obtained when varying a single parameter. $ARI_{best_r}$ represents the average of the best accuracies obtained when parameters are randomly selected.}
\centering
%\scriptsize
\resizebox{1.0\textwidth}{!}{\begin{minipage}{\textwidth}
\begin{tabular}{@{} |l|ccc|ccc|}
\cline{2-7}
\multicolumn{1}{l|}{} & \multicolumn{3}{c}{DB2C10F} & \multicolumn{3}{c|}{DB10C10F} \\
\hline
\multirow{2}{*}{\bf Algorithm} & 
\multirow{1}{*}{\bf$ARI_{def}$} & \multirow{1}{*}{\bf$ARI_{best_p}$} & \multirow{1}{*}{\bf $ARI_{best_r}$} & 
\multirow{1}{*}{\bf$ARI_{def}$} & \multirow{1}{*}{\bf$ARI_{best_p}$} & \multirow{1}{*}{\bf $ARI_{best_r}$} \\ 
& (\%) & (\%) & (\%) & (\%) & (\%) & (\%) \\

\hline
subspace     & 89.9 & 100.0  & 100.0  & 86.1 & 89.1 & 99.3 \\ 
EM     	  & 23.4 & 100.0 & 100.0 & 40.9 & 91.6 & 100.0 \\
spectral	  & 82.4 & 88.1  & 90.0  & 70.9 & 82.3 & 87.3  \\
clara	      & 53.0 & 61.0  & 70.0  & 51.9 & 57.6	& 68.0 \\
hcmodel      & 34.2 & 51.3  & 70.0  & 54.2 & 61.5	& 64.0 \\
k-means      & 36.6 & 53.5  & 60.0  & 52.0 & 57.8	& 60.0 \\
hierarchical & 0.0  & 43.7  & 80.0  & 13.8 & 61.5	& 68.0 \\	 
\hline
   
\end{tabular}
\end{minipage}}
\end{table*}
 
Other algorithms could be compared in future extensions of this work. An important aspect that could also be explored is to consider other statistical distributions for modeling the data. In addition, an analogous approach could be applied to semi-supervised classification. 

\section*{Acknowledgments}

M. M. Z. Rodriguez thanks CAPES for financial support. C. H. Comin thanks FAPESP (grant no. 15/18942-8) for financial support. D. R. Amancio thanks FAPESP (grant no. 14/20830-0 and 16/19069-9). O. M. Bruno gratefully acknowledges the financial support of CNPq (grant  no. 307797/2014-7) and FAPESP (grant no. 14/08026-1).
L. da F. Costa thanks CNPq (grant no. 307333/2013-2) and NAP-PRP-USP for sponsorship. This work has been supported also by FAPESP grant 11/50761-2.

\bibliography{amcapaper}

\newpage

\section*{\large Supplementary information for ``Clustering Algorithms: A Comparative Approach''}
\newpage

\renewcommand\thefigure{\thesection.\arabic{figure}}    
\setcounter{figure}{0}   
\renewcommand\thesection{S\arabic{section}}  
\setcounter{section}{0}   

\section{Description of the clustering algorithms' parameters}
In the following, we provide a brief description about the parameters of the clustering algorithms considered in the main text. We note that, since some algorithms do not have a default value for the number of clusters, in all cases we set this parameter as the number of clusters in the dataset.

\subsection*{k-means clustering}
The k-means algorithm used in the main text has the following parameters: 
 \begin{itemize}
 
  \item iter.max: integer, the maximum number of iterations. Default value: 10. 
  \item nstart:  integer, indicates how many random sets should be chosen. Default value: 1.
  \item algorithm: string, implementation of the k-means algorithm to use. Default value: ``Hartigan-Wong".
  \item centers: integer, number of clusters. Default value: number of clusters in dataset.
  
 \end{itemize}

 \subsection*{Clustering for large applications (clara)}
 The algorithm has the following parameters:
 \begin{itemize}
    \item metric: string, specifies the metric to be used for calculating dissimilarities between observations. 
   Default value: ``euclidean".
    \item sample: integer, number of samples to be drawn from the dataset. Default value: 5. 
    \item sampsize: integer, number of observations in each sample. sampsize should be larger than the number of clusters. Default value:  $\min(N, 40 + 2k)$, where $N$ is the number of objects.
    \item rngR: boolean, whether R’s random number generator should be used instead of the primitive clara. Default value: false.
    \item k: integer, the number of clusters. Default value: number of clusters in dataset.
\end{itemize}

\subsection*{Hierarchical clustering}
The hierarchical method has the following options: 
\begin{itemize}
 \item metric: string, metric to use for calculating distances between samples. Default value: ``Euclidean".
 \item method: string, clustering method to use. Default value: ``average".
 \item  par.method: integer, specifies the parameter for the dissimilarity calculation in some methods. Default value: 0.
 \end{itemize}
 
\subsection*{Expectation maximization (EM)}
The algorithm used for expectation maximization clusterization is provided by the \emph{mclust} package. Two routines of the package are used for applying the method:
 \begin{itemize}
  \item mstep: Maximization step in the EM algorithm for parametric Gaussian mixture models.
    \begin{itemize}
     \item z: string, conditional probability of the $i$-th observation belonging to the $k$-th component of the mixture. Default value: ``random".
     \item modelName: string, indicates the model to be used. Default value: ``VII".
    \end{itemize}

  \item estep: Implements the expectation step of EM algorithm for parameterized Gaussian mixture models. 
  \begin{itemize}
   \item modelName: string, indicates the model to be used. Default value: ``VII".
   \item parameters: List containing the mean, variance and mixing proportion for each component. These parameters are usually obtained in the expectation mstep. 
   \end{itemize}
 
 \end{itemize}

\subsection*{hcmodel clustering}
 Provided by the \emph{mclust} package. The hc routine employing the hcmodel has the following parameters: 
 \begin{itemize}
  \item modelName: string, indicates the model to be used. Default value: ``VII".
  \item use: string, specify what type of data/transformation should be used for model-based hierarchical clustering. Default value: ``VARS".
  \item G: integer, number of clusters. Default value: number of clusters in dataset.
 \end{itemize}

\subsection*{Spectral algorithm}
The routine specc of the \emph{kernlab} package has the following options:
\begin{itemize}
 \item centers: integer, number of clusters.  Default value: number of clusters in dataset.
 \item kernel: string, the kernel function used in computing the affinity matrix. Default value: ``rbfdot". The following options are available:\
 \begin{itemize}
    \item rbfdot: Radial Basis kernel function (``Gaussian").
    \item polydot: Polynomial kernel function.
    \item vanilladot: Linear kernel function.
    \item tanhdot Hyperbolic tangent kernel function.
    \item laplacedot Laplacian kernel function.
    \item besseldot: Bessel kernel function.
    \item anovadot: ANOVA RBF kernel function.
    \item splinedot: Spline kernel.
    \item stringdot: String kernel 
 \end{itemize}
 \item kpar: string, the  kernel parameter can also be set to a user defined function of class ``kernel" by passing the function name as an argument. Default value: ``automatic".
 \item  nystrom.sample: float, proportion of data to use when estimating sigma. Default value: $N_b/6$, where $N_b$ is the number of objects. 
 \item iter: integer, the maximum number of iterations. Default value: 200.
\end{itemize}

\subsection*{Subspace algorithm}
The routine used for subspace clustering, contained in the \emph{HDclassif} package, is called hddc (High Dimensional Data Clustering). It has the following parameters:
\begin{itemize}
 \item model: integer or string, 14 models can be used: 12 models with class specific orientation matrix and two models with common covariance matrix. Default value: ``akjbkQkdk" or 1. 

\item k: designates the number of clusters. The algorithm selects the result with the maximum BIC value. Default value: the selected k is in the default interval $(1,10]$. 
\item algo: string, the algorithm used for clustering. Can be either EM, CEM (Classification EM) or SEM (Stochastic EM). Default value: ``EM".
\item init: string, how to the initial class assignments are done. Default value: ``kmeans". Four initializations have been implemented:
\begin{itemize}
\item random: each observation is randomly assigned to a class.
\item kmeans: the initial class of each observation is provided by the k-means algorithm.
\item param:  initializes according to a multivariate normal distribution.
\item mini-em: the EM algorithm is run $m$ times for $nb$ iterations, the result with the highest likelihood is kept as the initialization of the algorithm.
\end{itemize}
\item mini.nb: integer, used when parameter init is ``mini-em". It is an array of length 2 containing $m$ and $nb$. Default value: (5, 10).
\end{itemize}

 \newpage
\section{Clustering performance obtained for random selection of parameters}

\begin{figure*}[h]
    \begin{center}
          \begin{subfigure}[][]{
	      {\includegraphics[width=0.45\columnwidth]{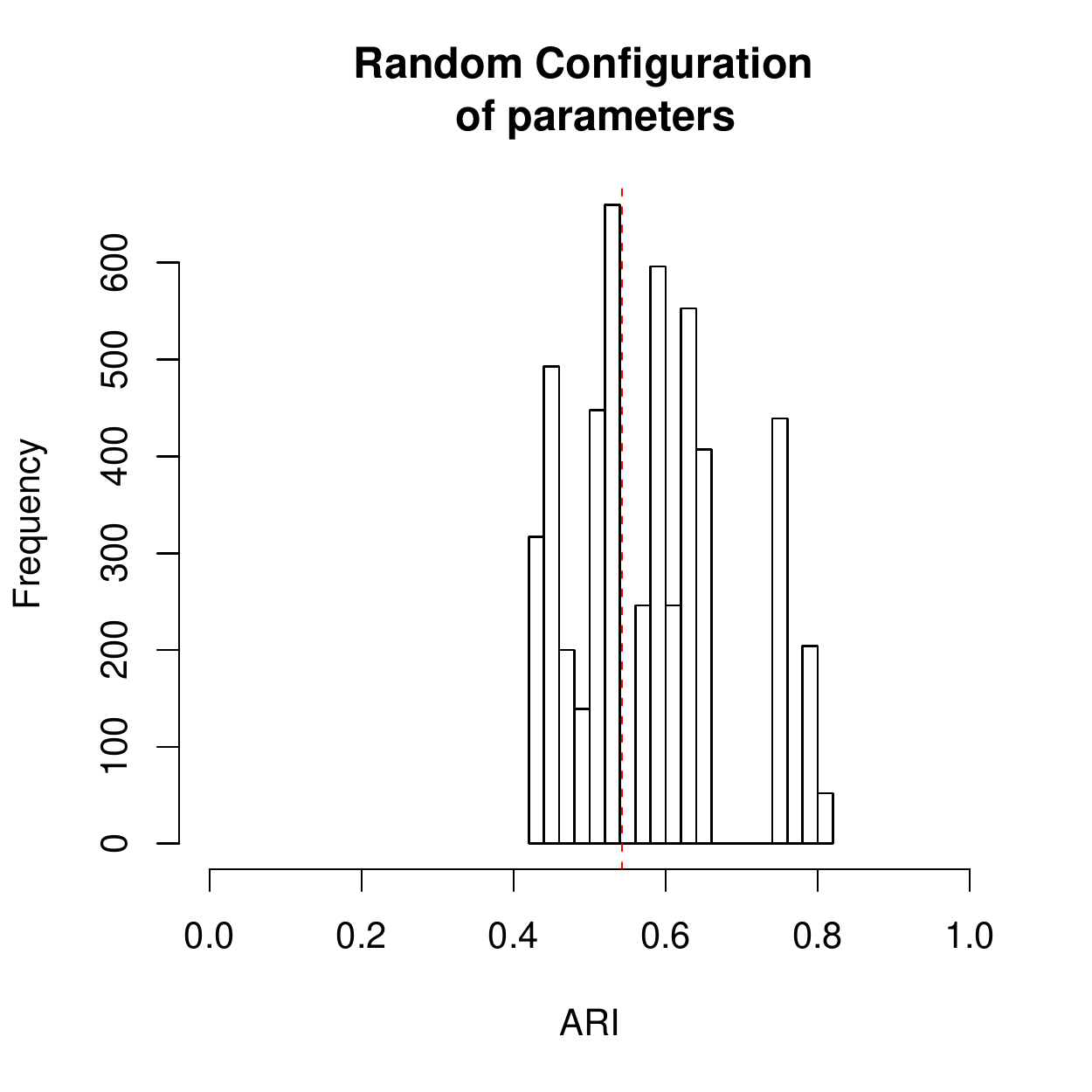}}\label{fig:Original_clusters}}
          \end{subfigure}
          \begin{subfigure}[][]{
	      {\includegraphics[width=0.45\columnwidth]{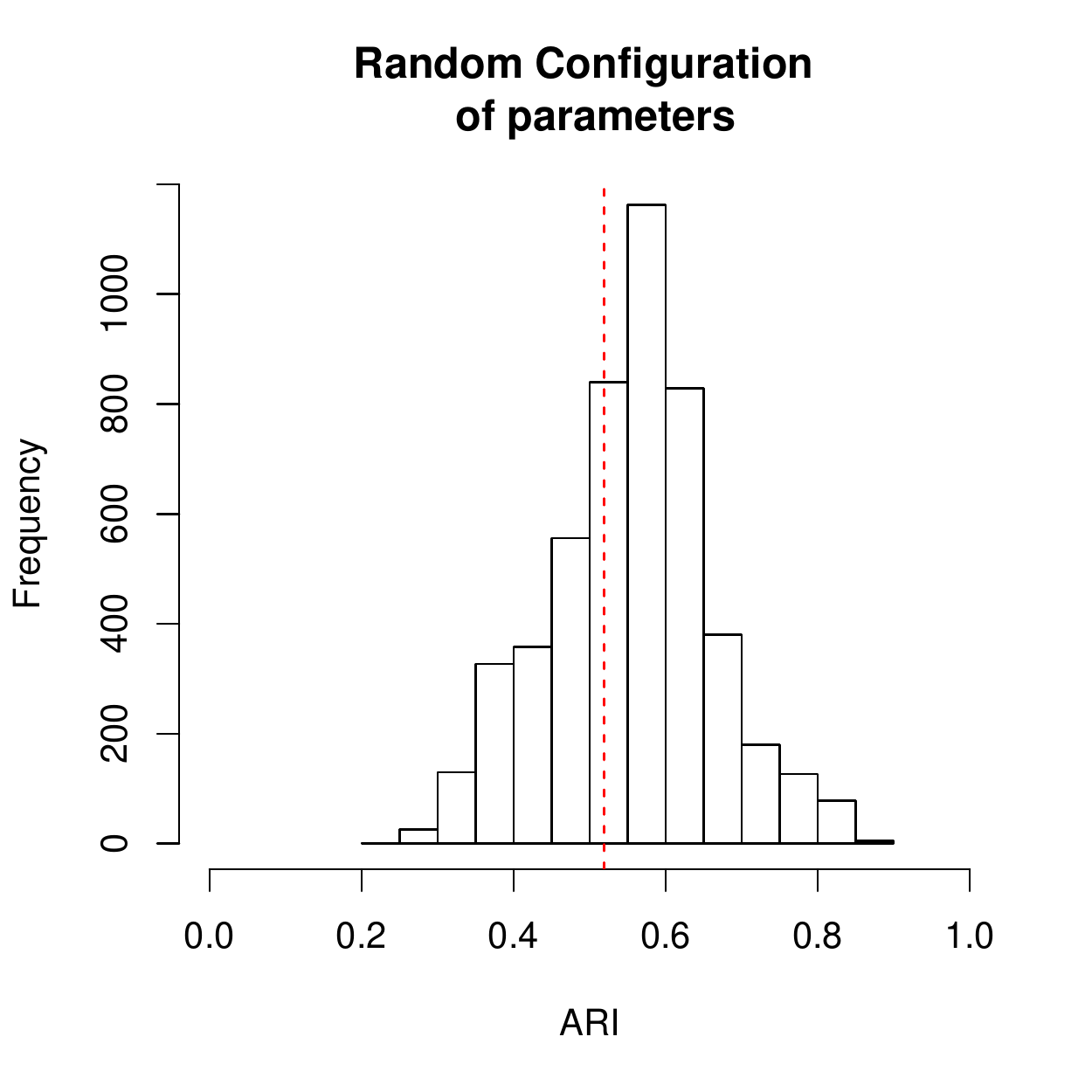}}\label{fig:Subspace_clusters}}
          \end{subfigure}
          \begin{subfigure}[][]{
	      {\includegraphics[width=0.45\columnwidth]{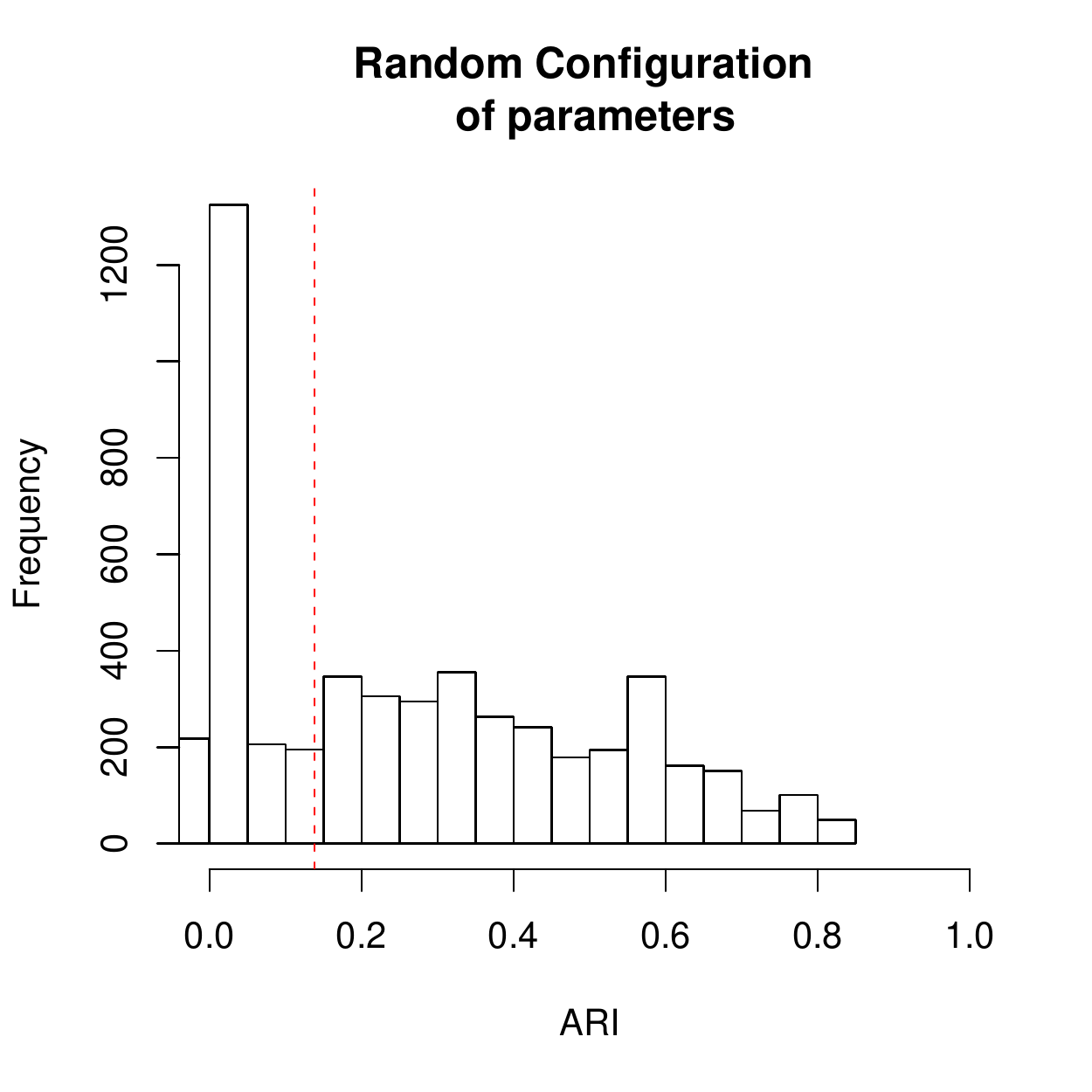}}\label{fig:Original_clusters}}
          \end{subfigure}
          \begin{subfigure}[][]{
	      {\includegraphics[width=0.45\columnwidth]{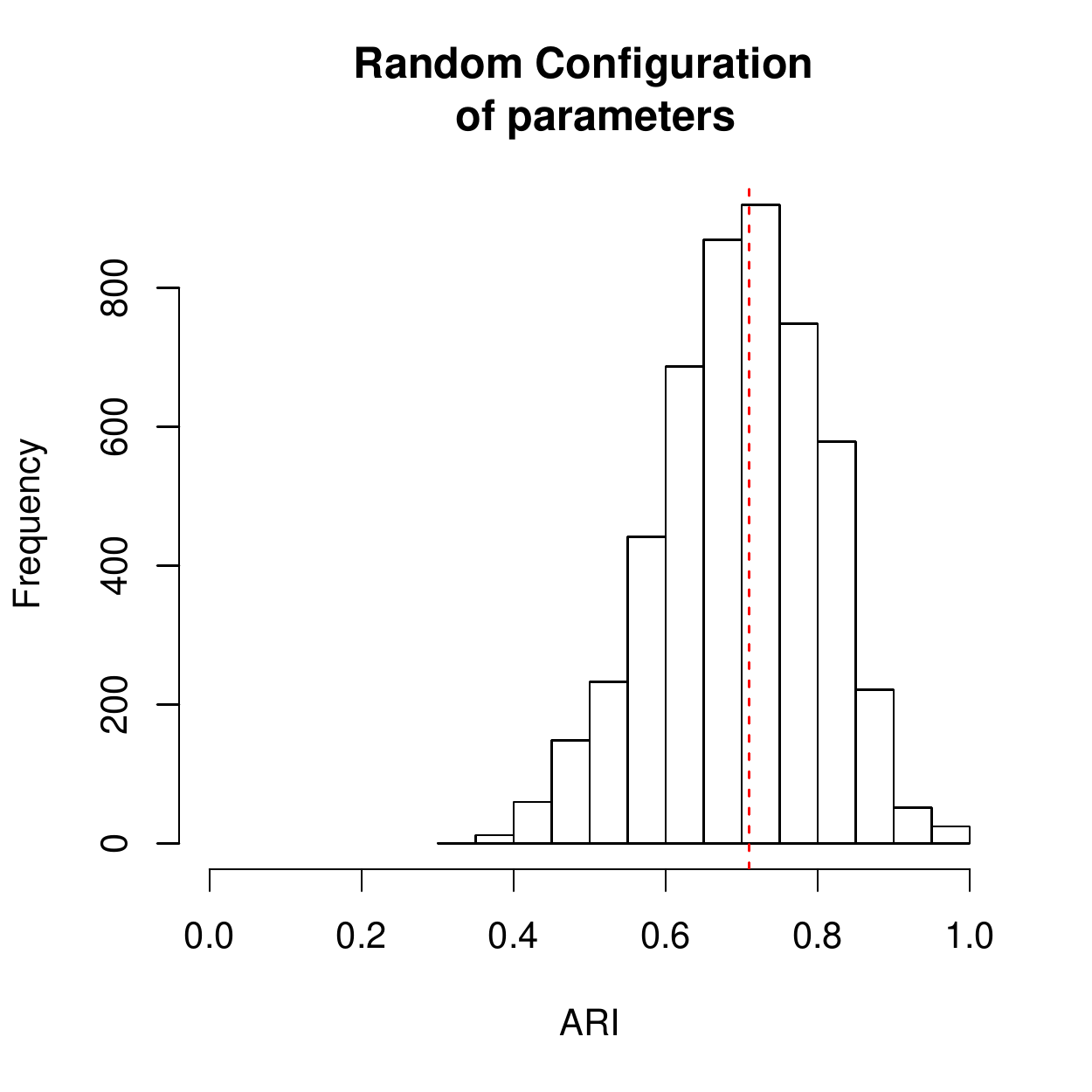}}\label{fig:Subspace_clusters}}
          \end{subfigure}
           \begin{subfigure}[][]{
	      {\includegraphics[width=0.45\columnwidth]{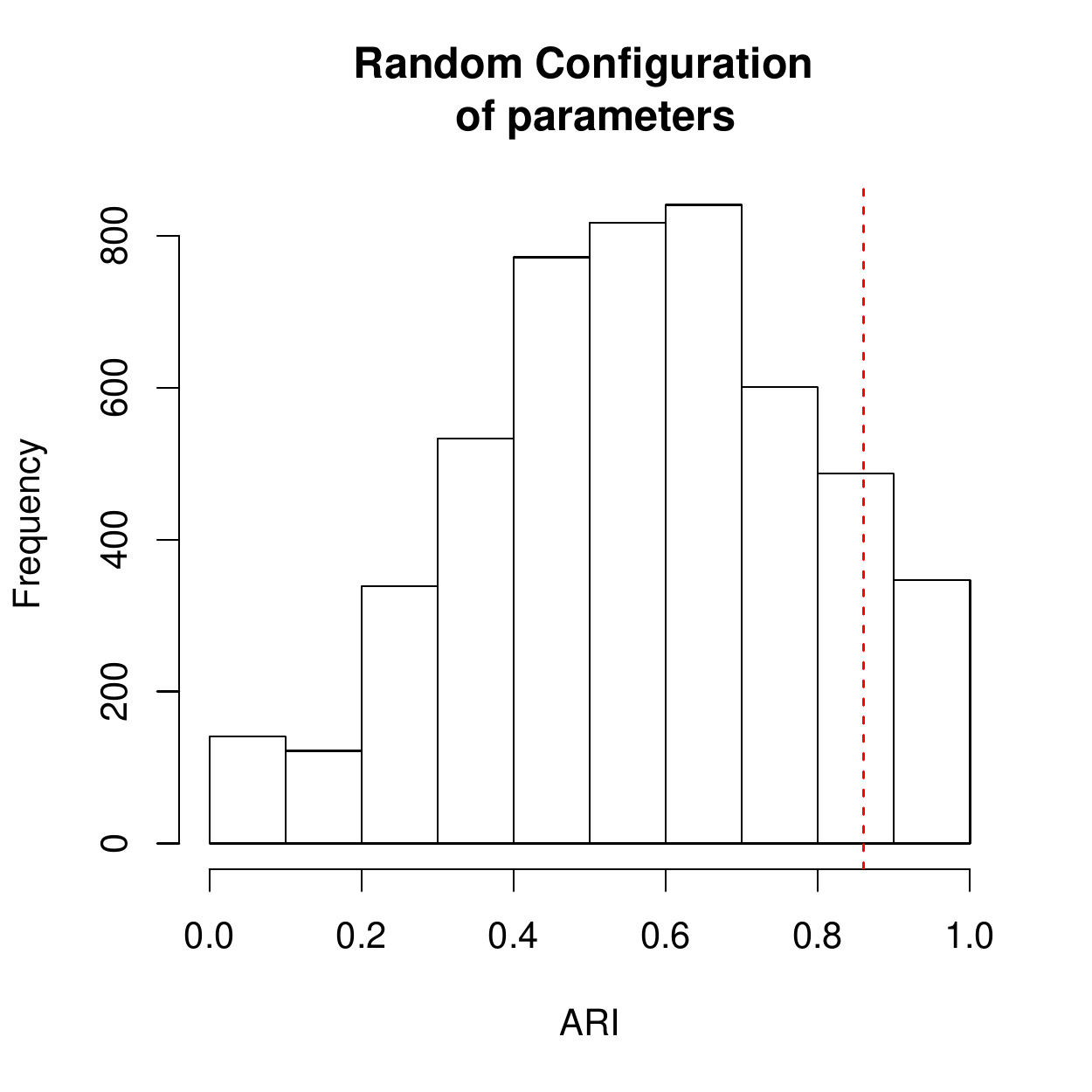}}\label{fig:Subspace_clusters}}
          \end{subfigure}
           \begin{subfigure}[][]{
	      {\includegraphics[width=0.45\columnwidth]{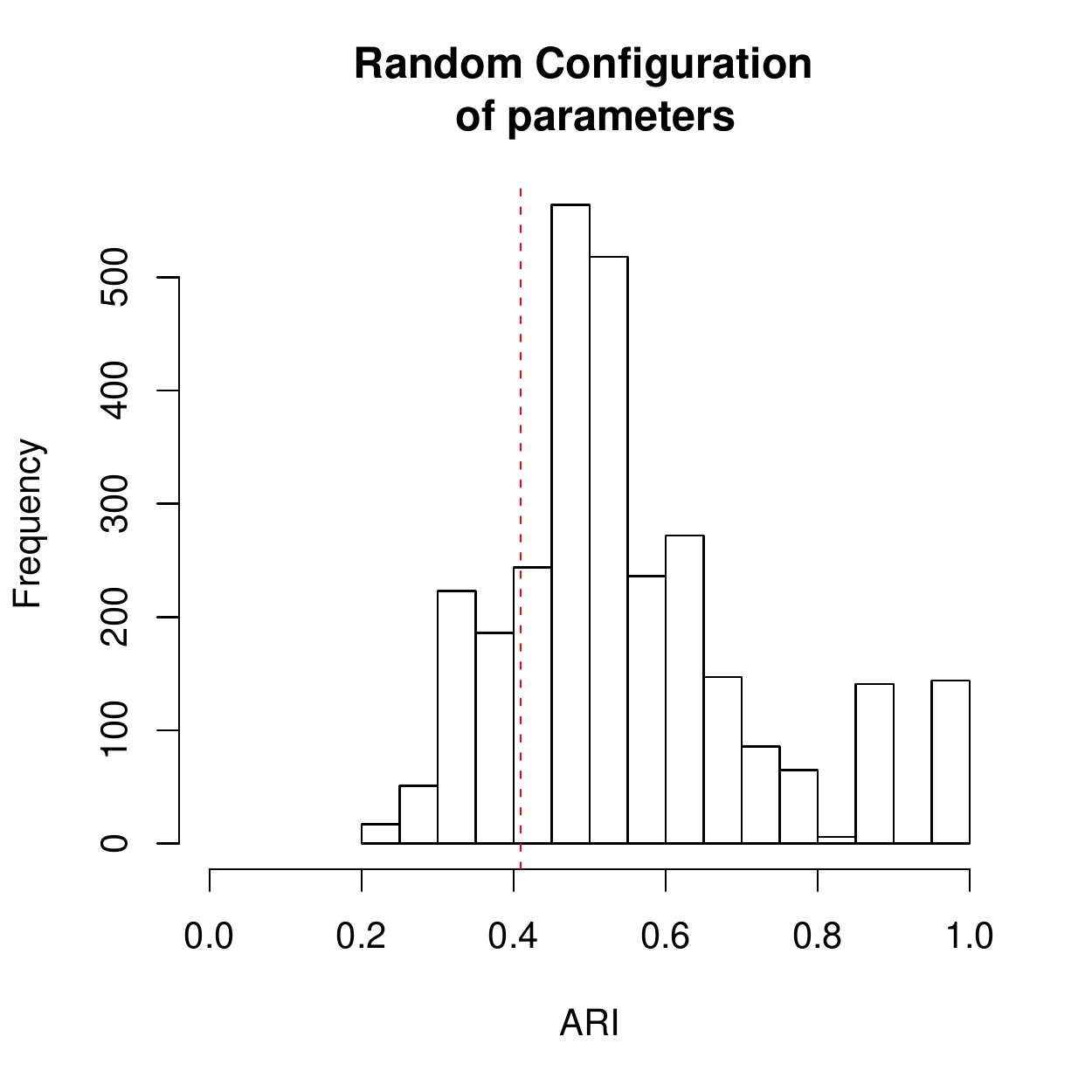}}\label{fig:Subspace_clusters}}
          \end{subfigure}

   \end{center}
\caption{{\bf Distribution of ARI values obtained for dataset DB10C10F using random selection of parameters.} The distributions correspond to the (a) \emph{hcmodel}, (b)  \emph{clara}, (c) \emph{hierarchical}, (d) \emph{spectral}, (e) \emph{Subspace} and (f) \emph{EM} methods. The red dashed line indicates the performance achieved when using the default parameters provided by the respective implementations of the algorithms.}
\label{fig:randomC10F10}
\end{figure*}

\begin{figure*}
    \begin{center}
          \begin{subfigure}[][]{
	      {\includegraphics[width=0.45\columnwidth]{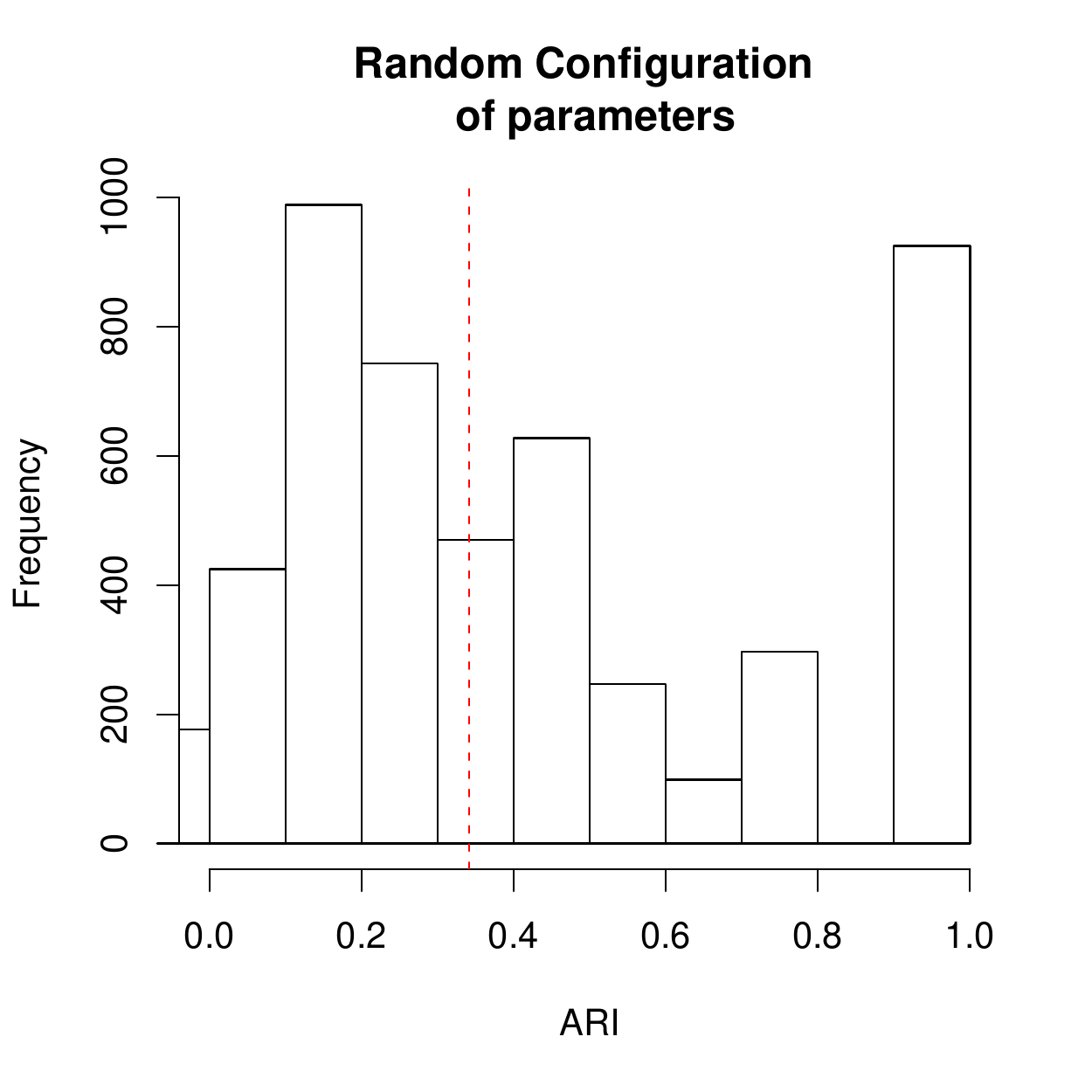}}\label{fig:Original_clusters}}
          \end{subfigure}
          \begin{subfigure}[][]{
	      {\includegraphics[width=0.45\columnwidth]{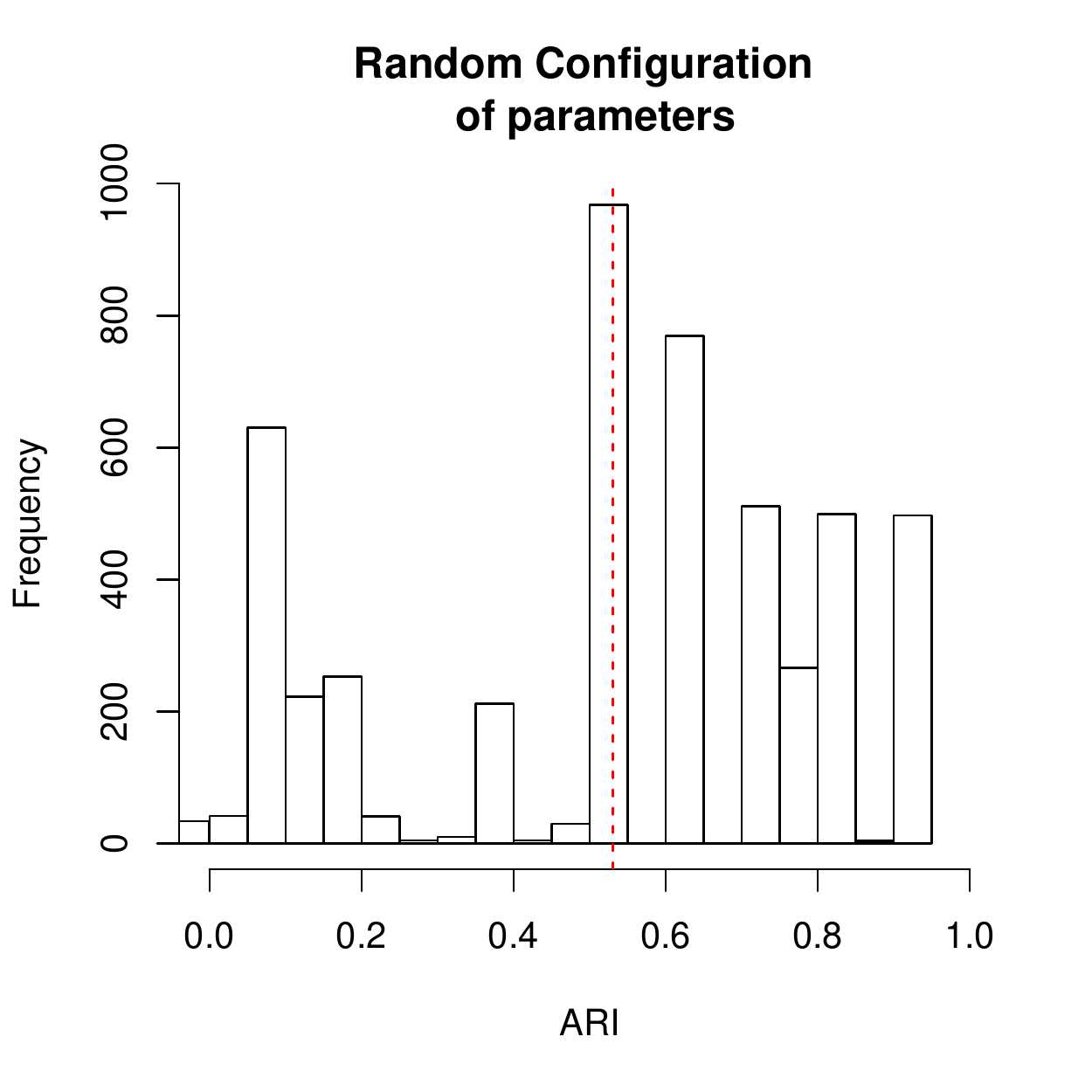}}\label{fig:Subspace_clusters}}
          \end{subfigure}
          \begin{subfigure}[][]{
	      {\includegraphics[width=0.45\columnwidth]{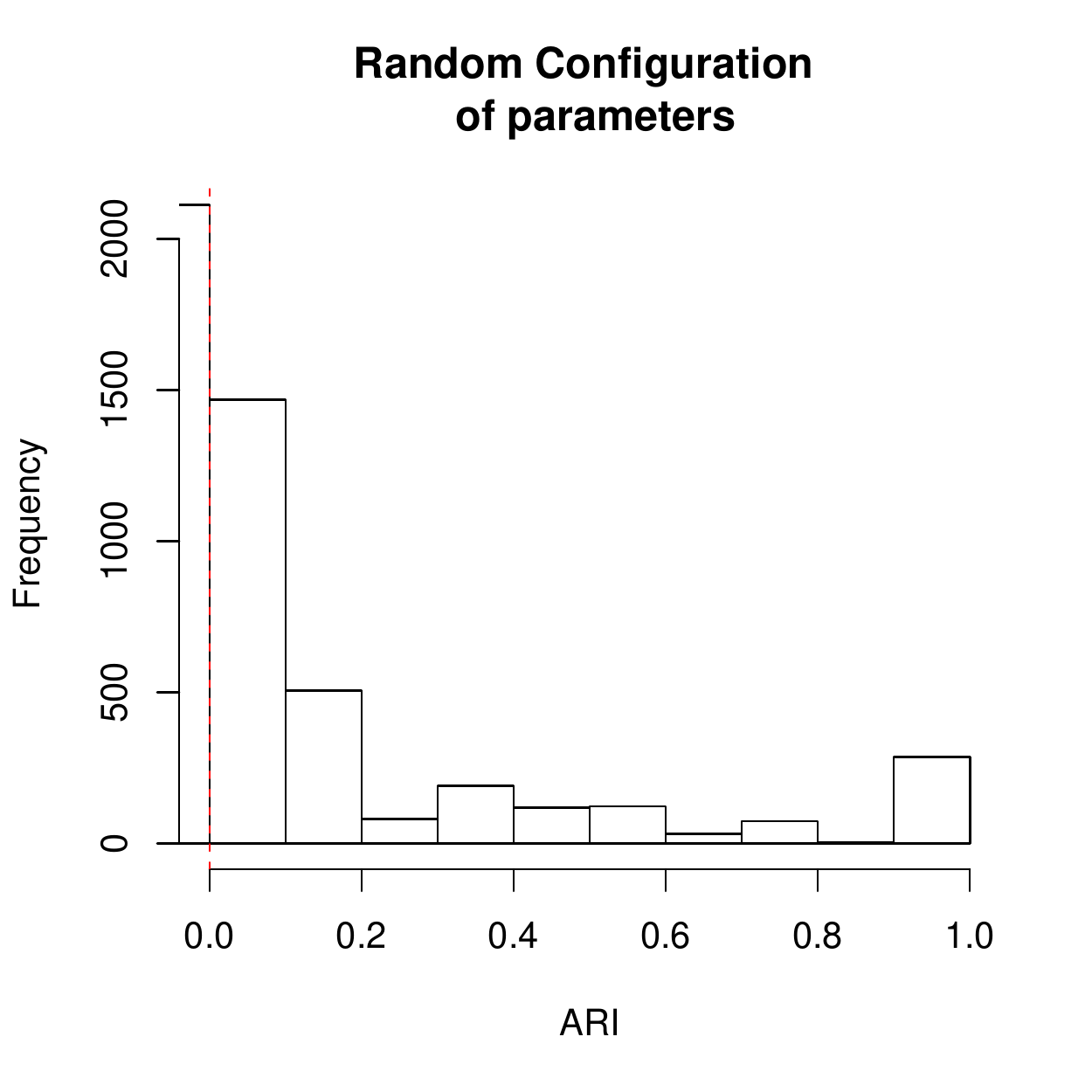}}\label{fig:Original_clusters}}
          \end{subfigure}
          \begin{subfigure}[][]{
	      {\includegraphics[width=0.45\columnwidth]{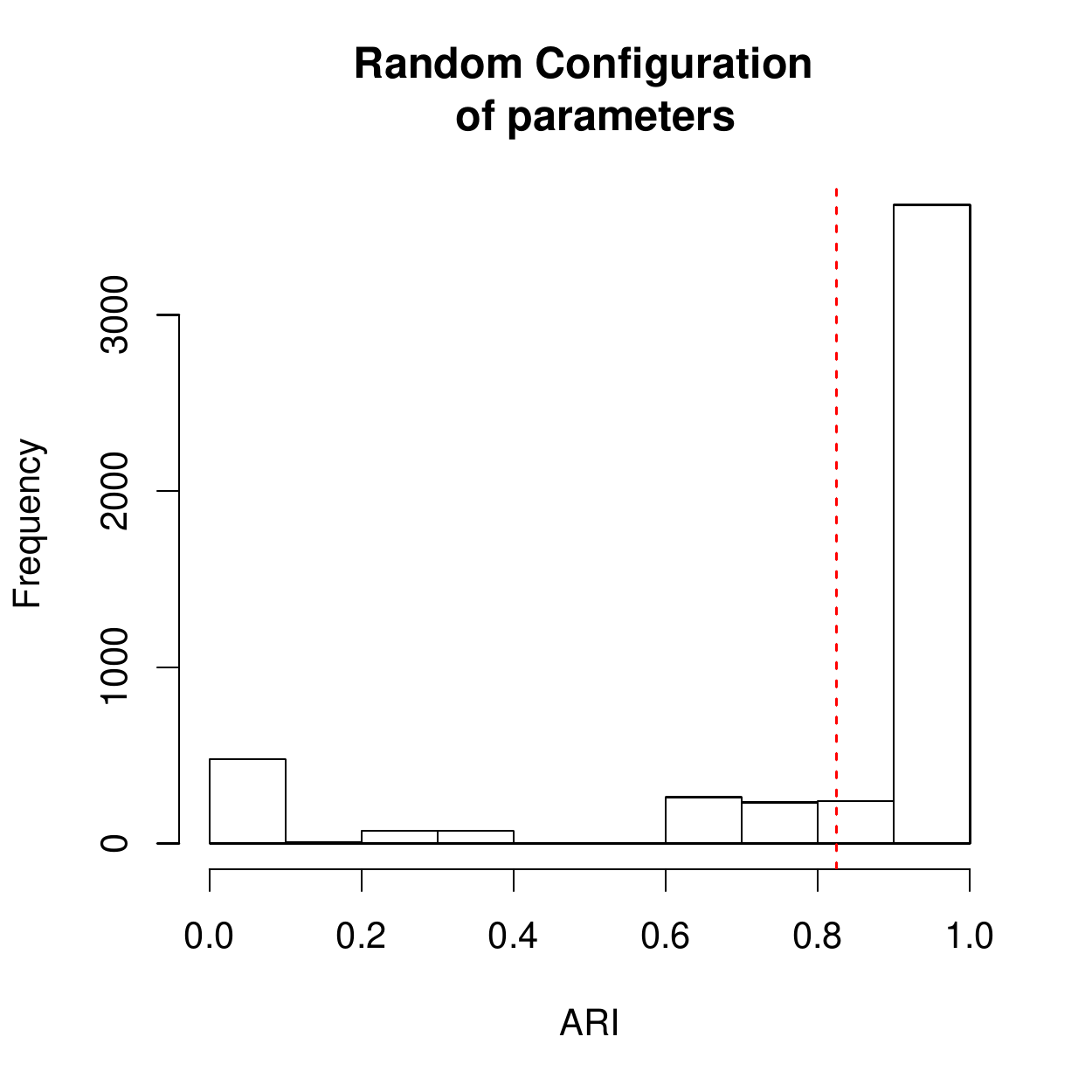}}\label{fig:Subspace_clusters}}
          \end{subfigure}
           \begin{subfigure}[][]{
	      {\includegraphics[width=0.45\columnwidth]{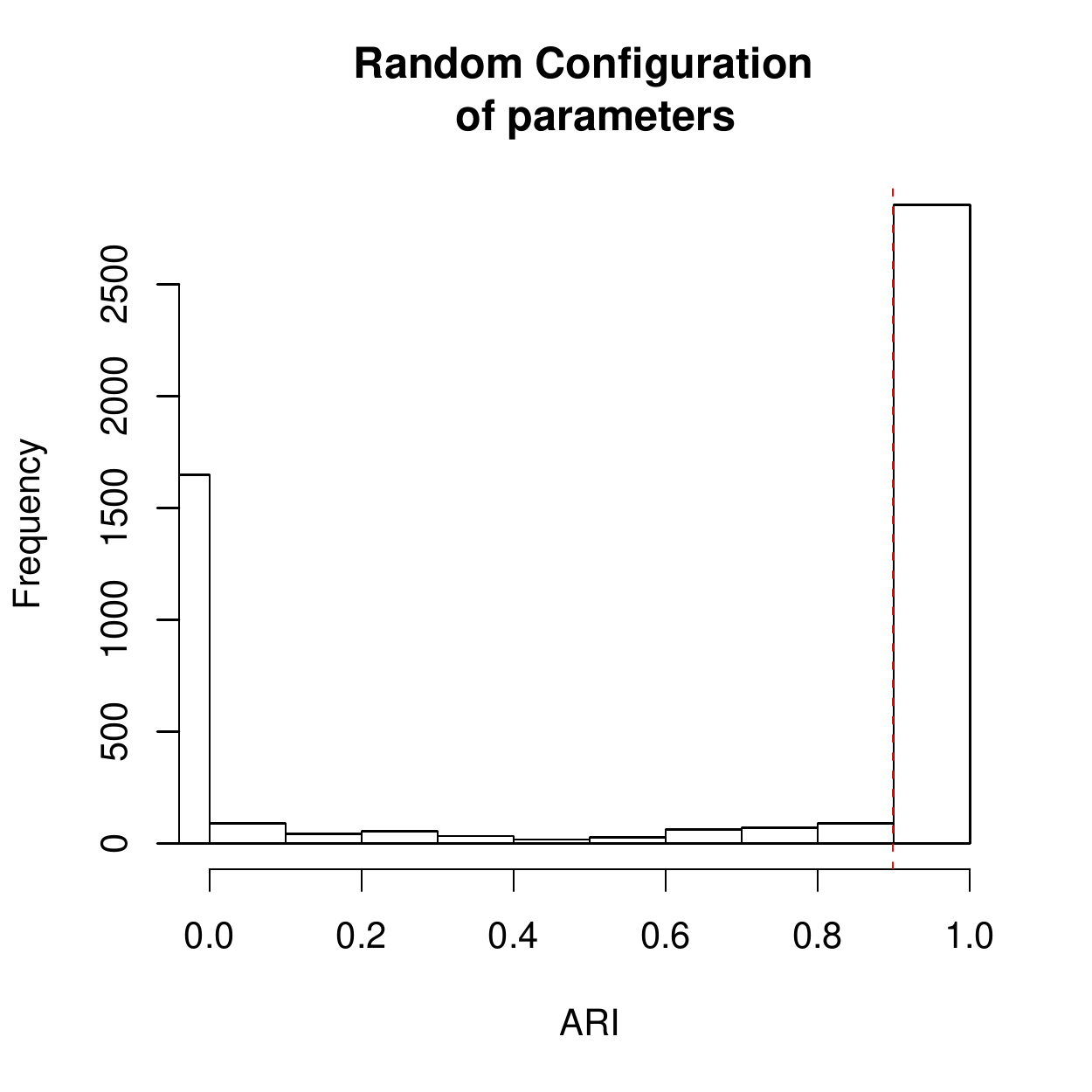}}\label{fig:Subspace_clusters}}
          \end{subfigure}
           \begin{subfigure}[][]{
	      {\includegraphics[width=0.45\columnwidth]{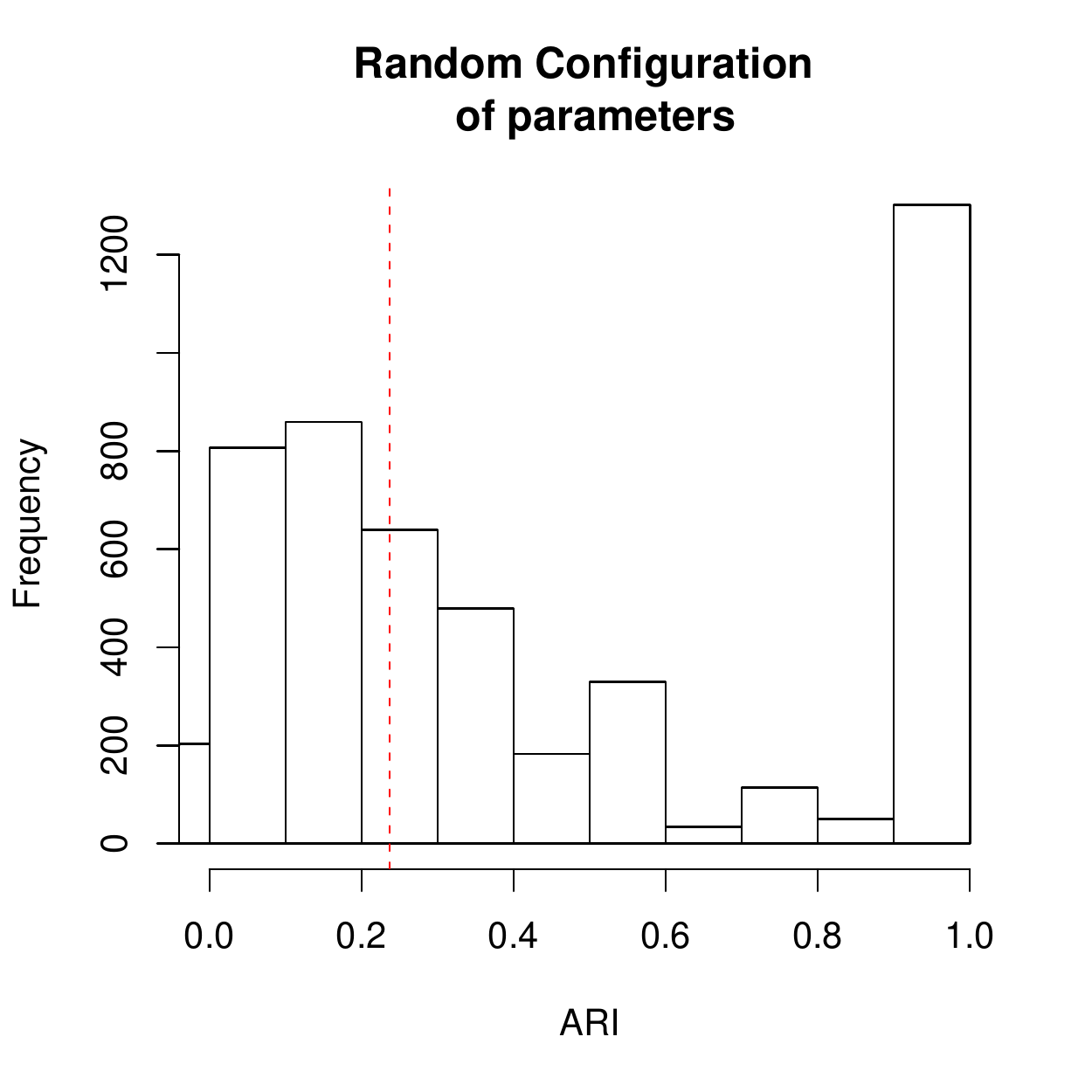}}\label{fig:Subspace_clusters}}
          \end{subfigure}

   \end{center}
\caption{{\bf Distribution of ARI values obtained for dataset DB2C10F using random selection of parameters.} The distributions correspond to the (a) \emph{hcmodel}, (b)  \emph{clara}, (c) \emph{hierarchical}, (d) \emph{spectral}, (e) \emph{Subspace} and (f) \emph{EM} methods. The red dashed line indicates the performance achieved when using the default parameters provided by the respective implementations of the algorithms.}
\label{fig:randomC2F10}
\end{figure*}

Figures~\ref{fig:randomC10F10} and~\ref{fig:randomC2F10} show the histograms of ARI values obtained for identifying the clusters of, respectively, datasets DB10C10F and DB2C10F using random selection of parameters. Each plot corresponds to a clustering method considered in the main text.

\end{document}